\documentclass[lettersize,journal]{IEEEtran}

\IEEEoverridecommandlockouts                              

\usepackage{epsfig} 
\usepackage{amsmath} 
\usepackage{amssymb}  

\usepackage{cite}

\usepackage{bm}
\usepackage{caption}
\usepackage{subcaption}
\usepackage[hidelinks]{hyperref}
\usepackage[shortlabels]{enumitem}

\usepackage{xcolor}

\renewcommand{\baselinestretch}{0.975}

\newcommand{\va}{\mathbf a}

\newcommand{\vf}{\mathbf f}

\newcommand{\vv}{\mathbf v}

\title{\LARGE \bf
	Impact-Aware Robotic Manipulation: \\ Quantifying the Sim-To-Real Gap for Velocity Jumps
}

\author{Jari van Steen, Daan Stokbroekx, Nathan van de Wouw, Alessandro Saccon
 \thanks{This work was partially supported by the Research Project I.AM. through the European Union H2020 program under GA 871899.}
 \thanks{Jari van Steen, Daan Stokbroekx, Nathan van de Wouw and Alessandro Saccon are with the Faculty of Mechanical Engineering, Eindhoven University of Technology, 5612 AE Eindhoven, The Netherlands (e-mail: j.j.v.steen@tue.nl, d.stokbroekx@student.tue.nl, n.v.d.wouw@tue.nl, a.saccon@tue.nl)}
}

\begin{document}

	\maketitle
	\thispagestyle{empty}
	\pagestyle{empty}

    \begin{abstract}
		
	Impact-aware robotic manipulation benefits from an accurate map from ante-impact to post-impact velocity signals to support, e.g., motion planning and control. This work proposes an approach to generate and experimentally validate such impact maps from simulations with a physics engine, allowing to model impact scenarios of arbitrarily large complexity. 
    This impact map captures the velocity jump assuming an instantaneous contact transition between rigid objects, neglecting the nearly instantaneous contact transition and impact-induced vibrations. 
    Feedback control, which is required for complex impact scenarios, will affect velocity signals when these vibrations are still active, making an evaluation solely based on velocity signals as in previous works unreliable. 
    Instead, the proposed validation approach uses the reference spreading control framework, which aims to reduce peaks and jumps in the control feedback signals by using a reference consistent with the rigid impact map together with a suitable control scheme. 
    Based on the key idea that selecting the correct rigid impact map in this reference spreading framework will minimize the net feedback signal, the rigid impact map is experimentally determined and compared with the impact map obtained from simulation, resulting in a 3.1\% average error between the post-impact velocity identified from simulations and from experiments. 
	\end{abstract}

\section{Introduction}\label{sec:introduction}

    Humans are naturally skilled in exploiting intentional impacts, as becomes apparent from many activities we perform daily, such as running, jumping, and catching a flying object. The growing field of impact-aware robotics focuses on having robots exploiting intentional impacts for a gain in performance, such as reduced execution time, for locomotion (i.e., jumping \cite{Zhang2020} or running \cite{Katz2019, Wensing2013}) and for manipulation (i.e. hitting \cite{Khurana2021}, catching \cite{Yan2024} or fast grabbing \cite{Bombile2022,Dehio2022}).
    Dealing with intentional impacts requires tackling challenges in different areas of research, including motion planning \cite{Zermane2024}, impact detection and classification \cite{Haddadin2017}, and control \cite{Wang2019}. 
    The availability of validated impact models is a key enabler to push beyond the state-of-the-art for each of these three research areas. 
    Examples for impact-aware motion planning include \cite{Kirner2024}, which uses an impact model to plan an impact motion that prevents immediate post-impact slippage, and \cite{Khurana2024}, which uses an impact model to execute a hitting motion with a given desired post-impact velocity. For impact classification, \cite{Proper2023} uses an impact model to distinguish between expected and unexpected impacts, with the aim of understanding whether a recovery action or safety stop should be triggered. In the field of impact-aware control, \cite{Yang2021} and \cite{Steen2024} use ante- and post-impact references compatible with an impact model in order to remove undesired jumps in the tracking error when transitioning from the ante- to the post-impact state. 

    \begin{figure}
		\centering
		\includegraphics[width=0.9\linewidth]{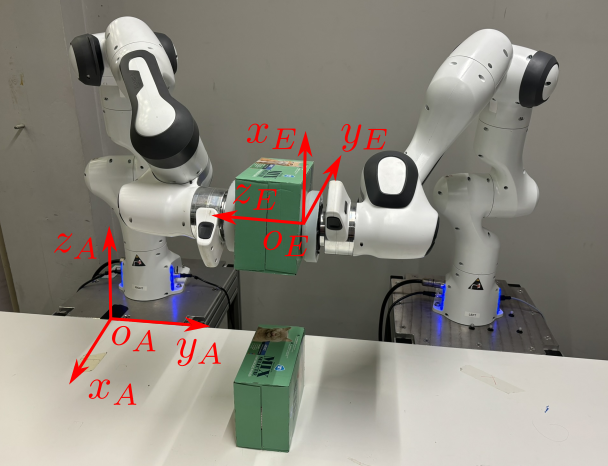}
		\caption{Depiction of the dual-arm robotic setup used throughout the validation approach presented in this work.}
		\label{fig:setup}
	\end{figure}	
    
    Impacts between robots and their environment can be modeled at different levels of abstraction as highlighted by \cite{Aouaj2021}, which distinguishes between:
    \begin{enumerate}[(a)]
    \item A compliant contact model with a flexible joint transmission model, taking into account the finite time duration of the impact (usually in the order of several milliseconds) and post-impact vibrations in the drivetrain.
    \item A nonsmooth contact model \cite{Brogliato2016, Glocker2006} with a flexible transmission model, assuming an instantaneous contact transition while still accounting for post-impact vibrations in the drivetrain.
    \item A \textbf{rigid impact map}, which combines a nonsmooth contact model with a rigid transmission model, assuming an instantaneous contact transition without oscillatory response in the drivetrain.
    \end{enumerate}

    It is noticeably challenging to accurately predict the transient velocity response shortly after the impact due to uncertainty of contact stiffness parameters for model (a), uncertainty of drivetrain stiffness and damping parameters for models (a) and (b), and other unmodeled vibrations in e.g. the environment or the robot structure. This challenge further increases when dealing with ideally simultaneous impacts, as is the case with jumping, dual-arm grabbing, or tasks with impacts between large surfaces. An unavoidable loss of simultaneity of the impacts can cause unplanned and unpredictable intermediate jumps in the transient velocity response before the velocities converge to a more predictable post-impact state once the impact sequence is finished \cite{Rijnen2019}. Especially in scenarios with simultaneous impacts, model (c) of the rigid impact map is therefore the only reliable and robust model one can have.

    A common rigid impact model, for example used in \cite{Kirner2024,Proper2023,Yang2021}, is the analytical impact model, which maps the ante-impact configuration and velocity into a post-impact velocity through a closed-form expression, while assuming the post-impact configuration remains equal due to the infinitesimal impact duration. However, for complex manipulation tasks, such as dual-arm grabbing with two 7 degree of freedom (DOF) robotic arms as perfomed in this work and shown in Figure~\ref{fig:setup}, an analytical impact model cannot be easily derived. 
    An alternative is presented in \cite{Steen2022b}, which computes the rigid impact map by simulating the impact event using a compliant contact model. Such a compliant contact model can, however, be computationally expensive for realistic use case scenarios involving many objects and robots due to the small timestep required for accurate results. Especially in use cases where impact maps are determined during robot operation, the resulting large computation time is undesired. Furthermore, compliant contact models demand challenging estimation of stiffness and damping contact parameters. Another alternative is presented in \cite{Steen2024}, where the rigid impact map is extracted from physical experiments. This, however, requires a training procedure and thus prevents the manipulation of previously unhandled objects or in unexplored impact configurations. 
    
    This work is centered around the idea of extracting the impact map from a physics engine based on a time-stepping method \cite{Acary2008}, which uses the theory of nonsmooth mechanics to efficiently evaluate the inelastic rigid impact map without requiring an estimation of contact stiffness and damping parameters. The extracted rigid impact map should, however, be experimentally validated before it can be trusted to be used reliably, especially considering prior work questioning the validity of rigid partially elastic impact models for simultaneous impact events \cite{Chatterjee1999}.

    Prior work in validation for robot-environment impacts includes \cite{Arias2024,Aouaj2021}, which focuses on a validation of an analytical impact map for single impacts. In particular, \cite{Arias2024} found that the most accurate impact map was one where not only the scaled motor inertia but also the low-level torque control loop effect is accounted for through the reflected motor inertia. In \cite{Acosta2022}, two physics engines (MuJoCo \& Drake) are validated against real-life data for a jumping humanoid. Both these physics engines, however, use compliant contact models. A third physics engine, Bullet, which does use a nonsmooth contact model description, was also compared in \cite{Acosta2022} for cube tossing. However, no simulations for robot-environment impacts were presented in Bullet due to its inability to account for the reflected motor inertia of the robot. In \cite{Jongeneel2023}, a validation of the physics engine AGX Dynamics, which uses a time-stepping approach with nonsmooth contact models, is performed for a box tossing application. However, no validation is performed for impacts between robots and their environment. 

    The main contribution of this work is, firstly, a demonstration that rigid impact maps for impacts between robots and their environment can be extracted from a physics engine that uses nonsmooth contact models. Secondly, we present an experimental validation to check the quality of these predictions for different reflected motor inertia models. 
    This validation will be provided for manipulation scenarios with simultaneous impacts for which, due to their complexity, computing an analytical contact model is undesirable or impossible, such as fast dual-arm grabbing of objects. Due to the presence of impact-induced oscillations in experiments that are not modeled in the rigid impact map, one cannot consider the velocity directly after an impact as representative for this impact map. Additionally, due to the challenging nature of the tasks, feedback control is required for successful task execution, causing the velocity to converge to whatever post-impact reference is provided before the impact-induced oscillations vanish. Hence, a validation of the impact map based solely on a comparison of the velocity jump between experiments and simulations, as in \cite{Aouaj2021,Arias2024}, is not possible.
    
    The validation procedure followed in this work makes use of the framework of reference spreading (RS) \cite{Saccon2014,Rijnen2015,Rijnen2017}. RS presents a way to reduce control input peaks and steps caused by so-called error peaking in traditional tracking control. This phenomenon concerns the problem that, under traditional tracking control, an unavoidable mismatch between the actual impact time and the predicted impact time will cause peaks in the velocity tracking error, and in the control inputs as a result \cite{Biemond2013,Forni2013,Leine2008}. By combining 1) an ante- and post-impact reference formulation that matches the rigid impact map with 2) an extension of these references past the nominal impact time and 3) a switching control policy based on impact detection, input peaks and steps are removed, minimizing control feedback during and after the impact sequence. The introduction of an interim-impact mode has additionally extended RS to be effective in the presence of simultaneous impacts as has been demonstrated in numerical simulation studies \cite{Rijnen2019,Steen2022} and in experimental work \cite{Steen2024}. 
    
    This paper aims to validate rigid impact maps by experimentally validating whether using the post-impact velocity prediction obtained from a simulation in an RS-based control framework causes the minimal feedback control effort, compared to using artificially different post-impact velocity predictions. 
    This analysis is based on the theory that, if the post-impact velocity prediction is correct, the post-impact reference velocity in the RS control framework will match the actual velocity. This in turn results in a control effort that is dominated by feedforward, and contains little feedback due to the low velocity tracking error. 
    By evaluating the integral of the feedback signal during and after the impact sequence for the different post-impact velocity predictions, an indirect experimental estimation of the optimal post-impact velocity is obtained, which is then compared against the predicted post-impact velocity to assess the validity of the physics engine. 
    We dedicate special attention to the effect that the motor inertia has on the rigid impact map by comparing the experimentally identified optimal post-impact velocity against the predicted post-impact velocities obtained through simulations using different motor inertia models, confirming the conclusions of \cite{Arias2024} also for complex simultaneous impact tasks.
    

    In the remainder of this paper, we introduce in Section \ref{sec:eom} the equations of motion of the robots used in the experimental evaluation, followed in Section \ref{sec:simulation} by an overview of the simulation framework used to generate the rigid impact map. In Section \ref{sec:refspread}, we present the RS-based control strategy used in the validation approach with details on the reference generation procedure and the actual control approach. We describe the methodology regarding the validation of the rigid impact map in more detail in Section \ref{sec:validation}, followed by the experimental results in Section \ref{sec:experiments} and the conclusion and recommendations for future research in Section \ref{sec:conclusion}.






\section{Robot equations of motion}\label{sec:eom}

    We will demonstrate the proposed rigid impact map validation approach for a single and dual-arm robotic manipulation use case, using the Franka Robotics dual-arm setup shown in Figure~\ref{fig:setup}. The robots are equipped with custom end-effectors with silicon cover \cite{Steen2024} preventing impact-induced safety limit violations, whose pose is parameterized by the coordinate frame $E$. Using a shortened variant of the notation of \cite{Traversaro2019}, we denote the position and rotation of the end effector frame $E$ with respect to the inertial frame $A$ as $\bm p := {}^A \bm p_{E}$ and $\bm R := {}^A \bm R_{E}$, respectively. We denote its twist as $\vv = \left[\bm v, \bm \omega\right]  := {}^{E[A]} \vv_{A,E}$, and its acceleration as $\va = \left[\bm a, \bm \alpha\right]  := {}^{E[A]} \va_{A,E}$ with linear and angular velocity, and linear and angular accelerations $\bm v, \bm \omega, \bm a$ and $\bm \alpha$, respectively.

	Each robot contains 7 actuated joints, whose displacements are represented by the joint displacement vector $\bm q \in \mathbb{R}^7$. The end effector twists and accelerations can be expressed in terms of the joint velocities and accelerations as
	\begin{equation}
	\vv = \bm J(\bm q) \dot{\bm{q}},
	\end{equation}
        \begin{equation}
	\va = \bm J(\bm q) \ddot{\bm{q}} + \dot{\bm J}(\bm q, \dot{\bm q}) \dot{\bm{q}} 
	\end{equation}
	with geometric Jacobian $\bm J(\bm q) :=  {}^{E[A]}\bm J_{A,E}(\bm q)$. While we are interested in a rigid robot model for the estimation of the post-impact velocity, in reality, there is a finite stiffness in the joint transmission of the robots. This leads to the well-known flexible joint model \cite{Albu2007} with equations of motion given by
    \begin{equation}\label{eq:EOM_link}
    \bm M(\bm q) \ddot{\bm q} + \bm h(\bm q,\dot{\bm q}) = \bm \tau_\text{meas} + \bm D \bm K^{-1} \dot{\bm \tau}_\text{meas} + \bm J^T(\bm q)  \mathbf{f}_c, 
    \end{equation}
    \begin{equation}\label{eq:EOM_motor}
    \bm B_\rho \ddot{\theta} + \bm \tau_\text{meas} + \bm D \bm K^{-1} \dot{\bm \tau}_\text{meas} = \bm \tau_{act} + \bm \tau_{\text{fric},m},
    \end{equation}
    \begin{equation}
    \bm \tau_\text{meas} = \bm K(\bm \theta - \bm q)
    \end{equation} 
    with link mass matrix $\bm M(\bm q)$, vector of gravity, centrifugal and Coriolis terms $\bm h(\bm q,\dot{\bm q})$, transmission joint torques $\bm \tau_\text{meas}$, motor side joint position $\bm \theta$ (as opposed to link side joint position $\bm q$), joint motor friction torques $\bm \tau_{\text{fric},m}$, contact wrench $\vf_c$, apparent motor inertia $\bm B_\rho$, actuation torque $\bm \tau_\text{act}$, and joint transmission stiffness and damping matrices $\bm K$ and $\bm D$ respectively. From \cite{Arias2024}, it is known that, when torque control is applied on the Franka robots, the inertia shaping low-level torque control law presented in \cite{Albu2007} given by 
    \begin{equation}\label{eq:torque_control_law}
        \bm \tau_\text{act} = \bm B_\rho \bm B_\theta^{-1} \bm \tau_d + (\bm I - \bm B_\rho \bm B_\theta^{-1})\left(\bm \tau_\text{meas} + \bm D \bm K^{-1} \dot{\bm \tau}_\text{meas}\right)
    \end{equation} 
    runs at 4 kHz, with desired joint torque $\bm \tau_d$ that is sent as input from a high-level controller, and desired reduced motor inertia matrix $\bm B_\theta$. Substitution of \eqref{eq:torque_control_law} in \eqref{eq:EOM_motor} results in the modified motor side equations of motion
        \begin{equation}\label{eq:EOM_motor_mod}
    \bm B_\theta \ddot{\bm \theta} + \bm \tau_\text{meas} + \bm D \bm K^{-1} \dot{\bm \tau}_\text{meas} = \bm \tau_{d} + \bm \tau_{\text{fric}}
    \end{equation}
    with reduced motor friction
    \begin{equation}
        \bm \tau_\text{fric} = (\bm B_\theta \bm B_\rho^{-1}) \bm \tau_\text{fric,m}.
    \end{equation}
    Following \cite{Ott2002,Arias2024}, we can substitute \eqref{eq:EOM_link} into \eqref{eq:EOM_motor_mod}, resulting in
    \begin{equation}
        \bm M(\bm q)\ddot{\bm q} + \bm B_\theta\ddot{\bm \theta} + \bm h(\bm q,\dot{\bm q})= \bm \tau_d + \bm \tau_\text{fric} + \bm J^T(\bm q)  \mathbf{f}_c, 
    \end{equation}
    and use the assumption of infinite joint stiffness $\bm K$, implying $\bm q = \bm \theta$, to formulate rigid equations of motion of the robot as 
	\begin{equation}\label{eq:eom}
	\left(\bm M(\bm q) + \bm B_\theta\right)\ddot{\bm q} + \bm h(\bm q,\dot{\bm q})= \bm \tau_d + \bm \tau_\text{fric} + \bm J^T(\bm q)  \mathbf{f}_c. 
	\end{equation} 
	Hence, we can use knowledge of the low-level torque control scheme to formulate rigid equations of motion with reduced motor inertia $\bm B_\theta$. The rigid model of \eqref{eq:eom} will be used throughout the remainder of this work. While the exact values of $\bm B_\rho$ and $\bm B_\theta$ are confidential, an estimation of the relevant parameters with an error margin of 10\%  provided by Franka Robotics can be found in Table~\ref{tab:parameters_B}. The values for $\bm M(\bm q)$ and  $\bm h(\bm q,\dot{\bm q})$ are taken by using the robot parameters presented in \cite{Gaz2019}. For ease of notation, the explicit dependency on $\bm q$ (or $\dot{\bm q}$) is dropped for the remainder of the document.

    \begin{table}[]
        \caption{Table containing relevant motor inertia parameter values for the Franka robot.}
        \begin{tabular}{l|l}
        Parameter                                                                                       & Value                                  \\ \hline
        $\bm B_\rho$                                                               & $\text{diag}(0.6,0.6,0.45,0.45,0.2,0.2,0.2) \text{kg}{\cdot}\text{m}^2$                                   \\
        $\bm B_\theta$                                                               & $\text{diag}(0.2,0.2,0.15,0.15,0.067,0.067,0.067) \text{kg}{\cdot}\text{m}^2$
        \end{tabular}
        \label{tab:parameters_B}
    \end{table}

\section{Simulation framework}\label{sec:simulation}

\begin{figure}
		\centering
		\includegraphics[width=0.92\linewidth]{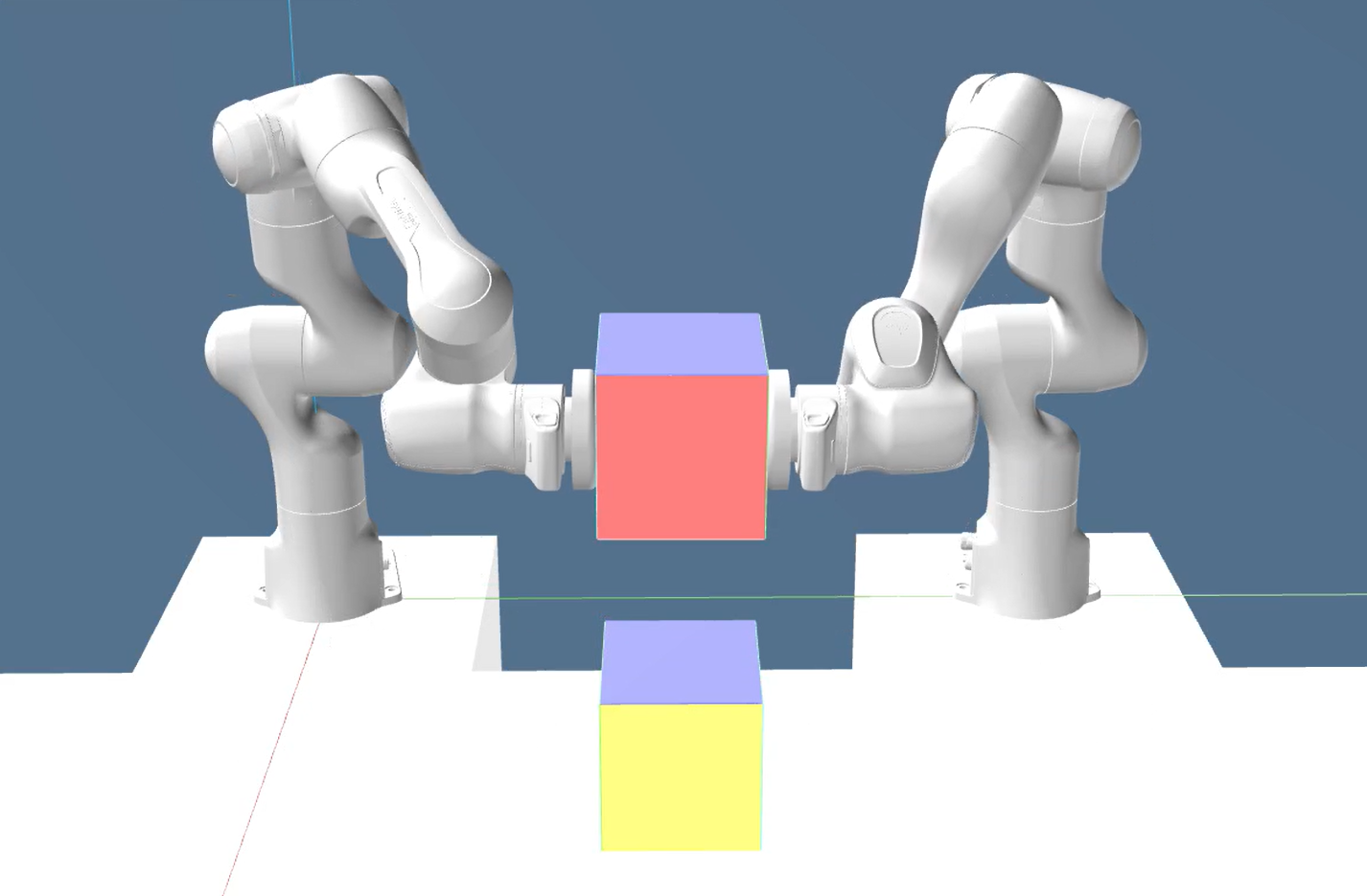}
		\caption{Visualization of a grabbing motion performed using the simulation framework.}
		\label{fig:simulation}
	\end{figure}

    The rigid impact map that is to be validated in this work is derived from simulations using the RACK\footnote{RACK: \href{https://gitlab.tue.nl/robotics-lab-public/glue-application}{https://gitlab.tue.nl/robotics-lab-public/glue-application}} simulation framework. 
    RACK is an open environment to describe a robot scene and the communication between a robot controller and a physics engine. It was developed in collaboration with Algoryx and CNRS-AIST JRL in the H2020 I.AM. project\footnote{H2020 I.AM. project: \href{https://i-am-project.eu/}{https://i-am-project.eu/}}. 
    Currently, the supported control software is the open source mc\_rtc\footnote{mc\_rtc: \href{https://jrl.cnrs.fr/mc_rtc/}{https://jrl.cnrs.fr/mc\_rtc/}} QP control framework. The supported physics engine is AGX Dynamics\footnote{AGX Dynamics: \href{https://www.algoryx.se/agx-dynamics/}{https://www.algoryx.se/agx-dynamics/}}, a well-established industrial physics simulator that has recently seen increased interest in academia \cite{Wiberg2022,Li2020,Styrud2022,Cao2023,Omer2024}, and for which academic licences are available. RACK can be used for setting up the simulation scene through a human-readable language called BRICK. Through a communication protocol called CLICK, simulations with synchronous robot control can be executed, and batch simulations can be performed through the ability to read and write data from and to HDF5 files. A screenshot of the virtual scene in the simulation framework for a grabbing motion is shown in Figure~\ref{fig:simulation}.

    AGX Dynamics uses a time-stepping approach \cite{Acary2008,Lacoursiere2007}, using the theory of nonsmooth mechanics \cite{Glocker2006,Brogliato2016}  to model unilateral contact, friction and impact. As highlighted in Section \ref{sec:introduction}, with time-stepping, impacts are resolved in a single simulation timestep without requiring contact stiffness or damping parameters. By only defining coefficients of restitution and friction, a rigid impact can thus be computed in this single timestep through evaluation of the velocity jump, which is the procedure followed in this work. All relevant friction coefficients in the simulation framework are identified through real-life experiments using a dynamometer, while the coefficients of restitution are all set to 0, assuming inelastic impacts. 
    Although not done for practical reasons, the expectation is that any other nonsmooth simulator would provide similar results, as long as the simulation can be initialized in a desired configuration and velocity.     
    In the remainder of this work, we will describe the approach to estimate the rigid impact map experimentally, and use that to validate the impact map estimated via the simulation framework.

\section{Control approach}\label{sec:refspread}

    Reference spreading (RS) is an impact-aware control approach that is central to the validation procedure as explained in Section \ref{sec:introduction}. The main idea of RS is to minimize jumps and peaks in the control input signals, in turn reducing the chance of destabilization, post-impact vibrations and hardware damage when tracking a trajectory with intentional impacts \cite{Steen2024}. In this work, a specific variant of the RS control scheme is formulated that aims to generate a zero net feedback during and after the impact, provided that the ante- and post-impact velocity references at the time of impact are matched through a rigid impact map that is consistent with the impact dynamics of the system. 
    The reference generation procedure is described in Section \ref{sec:refgen}, followed by the formulation of a Quadratic Programming (QP) based control approach in Section \ref{sec:control}. 

\subsection{Reference generation}\label{sec:refgen}

    To minimize the velocity tracking error and the corresponding input peaks and jumps, the two main ideas of the reference generation procedure in RS are as follows:
    \begin{enumerate}
        \item The velocity jump between the ante- and post-impact reference matches a rigid impact map, in this case provided by a physics engine.
        \item The ante- and post-impact references overlap around the nominal impact time, ensuring that a reference corresponding to the contact state is available, even if the impact occurs at a slightly different time than expected. This reduces input peaks that would otherwise negatively influence the results of the validation approach.
    \end{enumerate}
    Applying these principles, we will first discuss the procedure we used to formulate the ante-impact reference, followed by the formulation of a post-impact reference, tailored to our experiments using the Franka robot setup shown in Figure~\ref{fig:setup}. 

\subsubsection{Ante-impact reference}

    While path planning itself is not a contribution of this work and a range of path planning tools can be used in the RS framework, we will describe first what simple and effective approach was used in our experimental validation. This description is provided as a useful illustration of the two key principles behind RS, mentioned above. 
    
    Since the QP control approach uses tasks formulated in operational space \cite{Khatib1987}, a reference is generated for the end effector pose. Without loss of generality, the motions targeted in this paper maintain a constant end effector orientation $\bm R_{d}$. For the end effector position, a set of $N_a$ via points is defined, between which quintic trajectories \cite{Spong2020} are used to create an ante-impact reference $\bar{\bm p}^a_{d}(t)$ as 
    \begin{equation}
        \bar{\bm p}^a_{d}(t) = \sum_{k=0}^5 c^a_{j,k}t^k \ \text{for} \ t \in \left[T^a_{j-1},T^a_{j}\right)
    \end{equation}
    with transition times between segments defined as $T^a_j$ with $j\in \{0,N_a\}$. The bar sign over ${\bm p}^a_{d}$ indicates that the reference extends past the nominal impact time $T_\text{nom}$, following main idea 2) introduced at the beginning of this section. The parameters $c^a_{j,k}$ are uniquely determined through constraints 
    \begin{equation}\label{eq:ante_constr_refgen2}
    \bar{\bm p}^a_{d}(T^a_{j}) = \bm p^a_{j}, \ \dot{\bar{\bm p}}^a_{d}(T^a_{j}) = \bm v^a_{j}, \ \ddot{\bar{\bm p}}^a_{d}(T^a_{j}) = \bm a^a_{j}
    \end{equation}
    with user-defined parameters $\bm p^a_{j}, \bm v^a_{j}, \bm a^a_{j}$ for $j\in\{0,N_a\}$. This ensures continuity in position, velocity and acceleration between the reference segments. 

    To resolve the kinematic redundancy of the 7DOF robots used throughout this work, a polynomial $\bar{\xi}^a_{d}(t)$ is defined in similar fashion using quintic polynomials with via points. This reference prescribes the motion of the first joint as is done in \cite{Steen2024}.

\subsubsection{Post-impact reference}

    Similar to the ante-impact reference, the post-impact position reference $\bar{\bm p}^p_{d}(t)$ is created through quintic polynomials between $N_p$ via points, 
    resulting in 
    \begin{equation}
        \bar{\bm p}^p_{d}(t) = \sum_{k=0}^5 c^p_{j,k}t^k \ \text{for} \ t \in \left(T^p_{j-1},T^p_{j}\right]
    \end{equation}
    with constraints similar to \eqref{eq:ante_constr_refgen2} for parameters $\bm p^p_{j}, \bm v^p_{j}, \bm a^p_{j}$ and transition times $T^p_j$ with $j\in \{0,N_p\}$. At the nominal impact time $T_\text{nom}$, the ante- and post-impact references coincide, i.e.
    \begin{equation}\label{eq:pos_refs_imp}
        \bar{\bm p}^a_{d}(T_\text{nom}) = \bar{\bm p}^p_{d}(T_\text{nom}).
    \end{equation}
    To comply with main idea 1) highlighted at the beginning of this section, the velocity jump between the the ante-impact velocity reference $\dot{\bar{\bm p}}_d^a(T_\text{nom})$ and the post-impact velocity reference at the nominal impact time $\dot{\bar{\bm p}}^p_d(T_\text{nom})$ has to comply with the rigid impact map. This map is determined through the simulation framework highlighted in Section \ref{sec:simulation}, with the robots configured in the desired ante-impact configuration with initial end effector velocity $\dot{\bar{\bm p}}_d^a(T_\text{nom})$. The resulting post-impact end effector velocity is then used to determine $\dot{\bar{\bm p}}_d^p(T_\text{nom})$, ensuring that the velocity jump in the reference matches the rigid impact map. 

    The predicted post-impact velocity of joint 1 is also extracted from the simulations performed to compute the rigid impact map, and is used to define a post-impact reference $\bar{\xi}^p_{d}(t)$ similarly as done for $\bar{\bm p}^p_{d}(t)$. Given the nature of the flat end effectors seen in Figure~\ref{fig:setup} used for the experimental evaluation of this work, self-alignment between the end effector and the manipulated object is expected to occur, even if there is an initial angular velocity jump caused by the impact. Hence, the post-impact end effector orientation reference is given by the constant $\bm R_d$, identical to the ante-impact orientation reference.

\subsection{Control approach}\label{sec:control}

    Using the references generated in Section \ref{sec:refgen}, a control action is formulated with the aim of generating zero net feedback during and after the impact, provided that the ante- and post-impact velocity references at the time of impact comply with a a rigid impact map that is consistent with the impact dynamics of the system. An important feature to do so is using the RS framework to remove peaks from the feedback signals caused by impact-induced velocity jumps. In similar fashion to \cite{Steen2024}, this is done through the formulation of an ante-impact, an interim, and a post-impact control mode. 


\subsubsection{Ante-impact mode}\label{sec:control_ante}

    The aim of the ante-impact mode is to have the end effector follow the ante-impact position reference $\bar{\bm p}^a_{d}(t)$ with minimal feedback while maintaining a constant orientation $\bm R_d$, and resolve the kinematic redundancy of the 7DOF robots. Similar to \cite{Steen2024}, this is done through a QP control approach \cite{Bouyarmane2019,Salini2010}, computing the desired torques $\bm \tau_d$ by solving a QP optimization problem at every timestep. While not required for RS or the optimization approach, the use of QP control does enable the avoidance of, e.g., hardware limit violations and collision avoidance, and allows for multiple tasks to be enforced simultaneously, which is especially useful in more complex systems such as humanoids and quadrupeds. 
    The cost function is comprised of three tasks with corresponding error definitions. The position task error at acceleration level $\bm e^a_{p}$, aiming to follow the reference $\bar{\bm p}^a_{d}(t)$, is given by
    \begin{equation} \label{eq:e_pos_ante}
    \begin{aligned}
        \bm e^a_{p} := & \ \ddot{\bm p} - \ddot{\bar{\bm p}}^a_{d}(t) - 2\sqrt{k_p}\left( \dot{\bar{\bm p}}^a_{d}(t) - \dot{\bm p} \right) \\ & - k_p\left( {\bar{\bm p}}^a_{d}(t) - {\bm p} \right) - \bm a^a_\text{fric}(t)
        \end{aligned}
    \end{equation}
    with user-defined control gain $k_p$ and friction compensation term $\bm a^a_\text{fric}(t)$. 
    This feedforward friction compensation is particularly relevant for the validation approach, as we want to end up with no net input signals for the validation approach provided that the impact model is correct, hence it should be avoided that friction is compensated through feedback. 
    A pragmatic procedure for determining $\bm a^a_\text{fric}(t)$ is given in Appendix A. The orientation task error, aiming to maintain constant orientation $\bm R_d$, is given by
    \begin{equation} \label{eq:e_theta_ante}
        \bm e^a_{\theta} := \bm \alpha + 2\sqrt{k_\theta} \bm \omega + k_\theta(\log(\bm R^T{\bm R}_{d}(t)))^{\vee }
    \end{equation}
    with user-defined control gain $k_\theta$. Please note that the lack of acceleration and velocity feedforward is a consequence of the desired rotation being constant before and after the impact.  
    Finally, the posture task error, aiming to prescribe the motion of the first joint to resolve kinematic redundancy for the 7DOF robots, is given by
    \begin{equation}
         e^a_q :=  \ddot{\xi} - 2 \sqrt{k_q}(\dot{\bar{\xi}}^a_{d}(t) - \dot{\xi}) - k_q(\bar{\xi}^a_{d}(t) - \xi)
    \end{equation}
    with user-defined control gain $k_q$. 
    The QP for a single robot is then given by
    \begin{equation}\label{eq:QP_ante}
        \ddot{\bm q}^* = \underset{\ddot{\bm q}}{\operatorname{argmin}}  \left(w_{p}\|{\bm e^a_{p}}\|^2 + w_{\theta} \|{\bm e^a_{\theta}}\|^2 + w_{q} |e_{q}^a|^2 \right),
    \end{equation}
    subject to 
    \begin{equation}\label{eq:const_q}
        {\bm q}_\text{min} \leq \frac{1}{2}\ddot{\bm q} \Delta t^2 + \dot{\bm q}\Delta t + \bm q \leq {\bm q}_\text{max},
    \end{equation} 
    \begin{equation}\label{eq:const_dq}
        \dot{\bm q}_\text{min} \leq \ddot{\bm q} \Delta t + \dot{\bm q} \leq \dot{\bm q}_\text{max},
    \end{equation} 
    \begin{equation}\label{eq:const_tau}
        {\bm \tau}_\text{min} \leq (\bm M + \bm B_\theta)\ddot{\bm q} + \bm h  \leq {\bm \tau}_\text{max}
    \end{equation}     
    with control timestep $\Delta t$. The constraints \eqref{eq:const_q}-\eqref{eq:const_tau} are included not to exceed the upper and lower joint position, velocity and torque limits given by ${\bm q}_\text{max}, {\dot{\bm q}}_\text{max}$, ${\bm \tau}_\text{max}$ and ${\bm q}_\text{min}, {\dot{\bm q}}_\text{min}$, ${\bm \tau}_\text{min}$, respectively. The resulting desired acceleration $\ddot{\bm q}^*$ is then transformed into a desired torque $\bm \tau_d$ using the free space and frictionless equivalent of \eqref{eq:eom} as
    \begin{equation}\label{eq:eom_ante}
        \bm \tau_d = \left(\bm M + \bm B_\theta\right)\ddot{\bm q}^* + \bm h.  
    \end{equation} 
    The inclusion of $\bm B_\theta$ in \eqref{eq:eom_ante} is a novelty compared to \cite{Steen2024}, resulting in a substantial improvement of the tracking control performance. The resulting $\bm \tau_d$ is then sent as input for the low-level torque control loop \eqref{eq:torque_control_law}. 

\subsubsection{Interim mode}\label{sec:control_interim}


    Following the RS approach of \cite{Steen2024}, the interim mode becomes active after detection of the first impact and remains active for a set amount of time $\Delta t_\text{int}$. The aim of this mode is to reduce input peaks during and directly after the impact sequence, which can otherwise occur as a result of peaks in the velocity tracking error, for example caused by impact-induced vibrations. 
    The position task error in the interim mode is given by
    \begin{equation} \label{eq:e_pos_int}
        \begin{aligned}
        \bm e^\text{int}_{p} := & \ddot{\bm p} - \ddot{\bar{\bm p}}^p_{d}(t)    - 2\sqrt{k_p}\gamma(t)\left( \dot{\bar{\bm p}}^p_{d}(t) - \dot{\bm p} \right) \\ & - k_p \left( {\bar{\bm p}}^p_{d}(t) - {\bm p} \right) - \bm a^p_\text{fric}(t) - \bm a^p_f(t)
        \end{aligned}
    \end{equation}
    with post-impact friction compensation term $\bm a^p_\text{fric}(t)$ highlighted in Appendix A, external force compensation term $\bm a^p_f(t)$ clarified further down below (cf. \eqref{eq:a_p_f} in Section \ref{sec:contol_post}), and blending parameter $\gamma(t)$ as
    \begin{equation}
        \gamma(t) := \frac{t-T_\text{imp}}{\Delta t_\text{int}}
    \end{equation}
    with $T_\text{imp}$ defined as the detected impact time, which is when the interim mode is activated. 
    Due to the uncertainty of the contact state and the presence of impact-induced vibrations, velocity feedback is unreliable directly after an impact is detected. Hence, velocity feedback is initially set to 0, and slowly increases to the level set in the post-impact mode using $\gamma(t)$. A similar strategy with initial removal of velocity feedback is applied for the orientation task and posture task, such that the desired accelerations and torques can be computed as in \eqref{eq:QP_ante}-\eqref{eq:eom_ante}.

\subsubsection{Post-impact mode}\label{sec:contol_post}

    The QP in the post-impact mode is identical to the interim mode QP with $\gamma(t)=1$, having identical constraints \eqref{eq:const_q}-\eqref{eq:const_tau} as in the ante-impact and interim mode, and using \eqref{eq:eom_ante} to compute the desired torque. The post-impact position task error is thus given by 
    \begin{equation} \label{eq:e_pos_post}\begin{aligned}
        \bm e^p_{p} := & \ \ddot{\bm p} - \ddot{\bar{\bm p}}^p_{d}(t) - 2\sqrt{k_p}\left( \dot{\bar{\bm p}}^p_{d}(t) - \dot{\bm p} \right)  \\ & - k_p\left( {\bar{\bm p}}^p_{d}(t) - {\bm p} \right) - \bm a^p_\text{fric}(t) - \bm a^p_f(t).
        \end{aligned}
    \end{equation}
    The term $\bm a^p_f(t)$ is added with the aim of exerting a force on the manipulated object. The equivalent desired acceleration to exert a desired external wrench $\mathbf{f}_{d}$ is given by
    \begin{equation}\label{eq:a_p_f}
        \mathbf{a}^p_f = \begin{bmatrix} \bm a_f^p \\ \bm \alpha_p^f\end{bmatrix} = \bm \Lambda^{-1} \mathbf{f}_{d},
    \end{equation}
    with task-space inertia matrix $\bm \Lambda$ as
    \begin{equation}
        \bm \Lambda := \left(\bm J(\bm M + \bm B_\theta)^{-1} \bm J^T\right)^{-1},
    \end{equation}
    The desired wrench aims to apply a normal contact force and compensate for the external forces exerted on the end effectors through interaction with the object. These external forces depend on the use case, and can include gravity compensation forces or friction forces due to sliding, which can be modeled assuming known friction coefficients. On top of this, an orientation task enforcing the constant orientation $\bm R_d$ is added to the cost function together with a posture task to follow $\bar{\xi}^p_{d}(t)$. 

\section{Validation methodology}\label{sec:validation}

    This section will provide details on the core contribution of this work, which is a validation procedure for a rigid impact map for ideally simultaneous impacts. This impact map relates the ante-impact velocities to the post-impact velocities at the impact configuration, and is computed through a simulation-based approach, highlighted in Section \ref{sec:simulation}. The presence of impact-induced oscillations and a possible loss of simultaneity of the impacts in experimental conditions will cause an unpredictable transient response at each repetition, as highlighted in Section \ref{sec:introduction}. The necessity to apply feedback control during and after the impact sequence for realistic impact scenarios such as the tasks considered in this work will influence the post-impact velocities during this transient response. This implies that an open-loop validation based solely on velocity signals, as proposed in \cite{Arias2024} for simpler robot-environment impact experiments with no objects, is not suitable.

    \begin{figure*}
		\centering
		\begin{subfigure}[b]{0.324\textwidth}
			\centering
			\includegraphics[width=\textwidth]{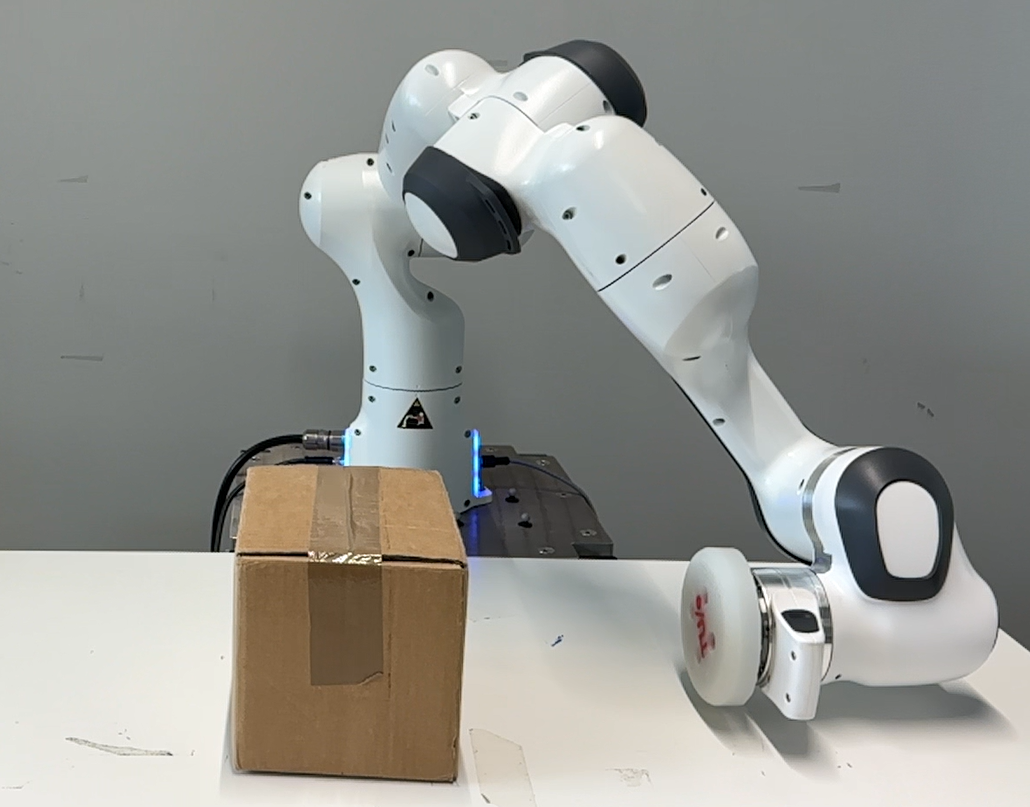}
			\caption{Ante-impact state}
			\label{fig:Push_sequence_1}
		\end{subfigure}
		\begin{subfigure}[b]{0.2583\textwidth}
			\centering
			\includegraphics[width=\textwidth]{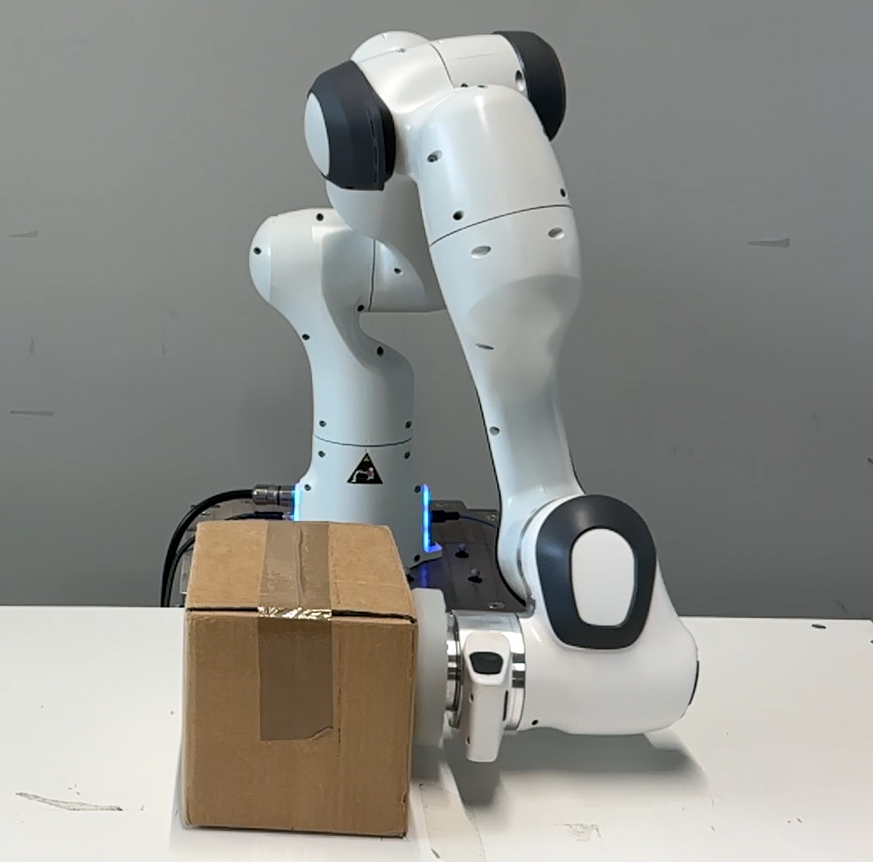}
			\caption{Impact configuration}
			\label{fig:Push_sequence_2}
		\end{subfigure}
		\begin{subfigure}[b]{0.2952\textwidth}
			\centering
			\includegraphics[width=\textwidth]{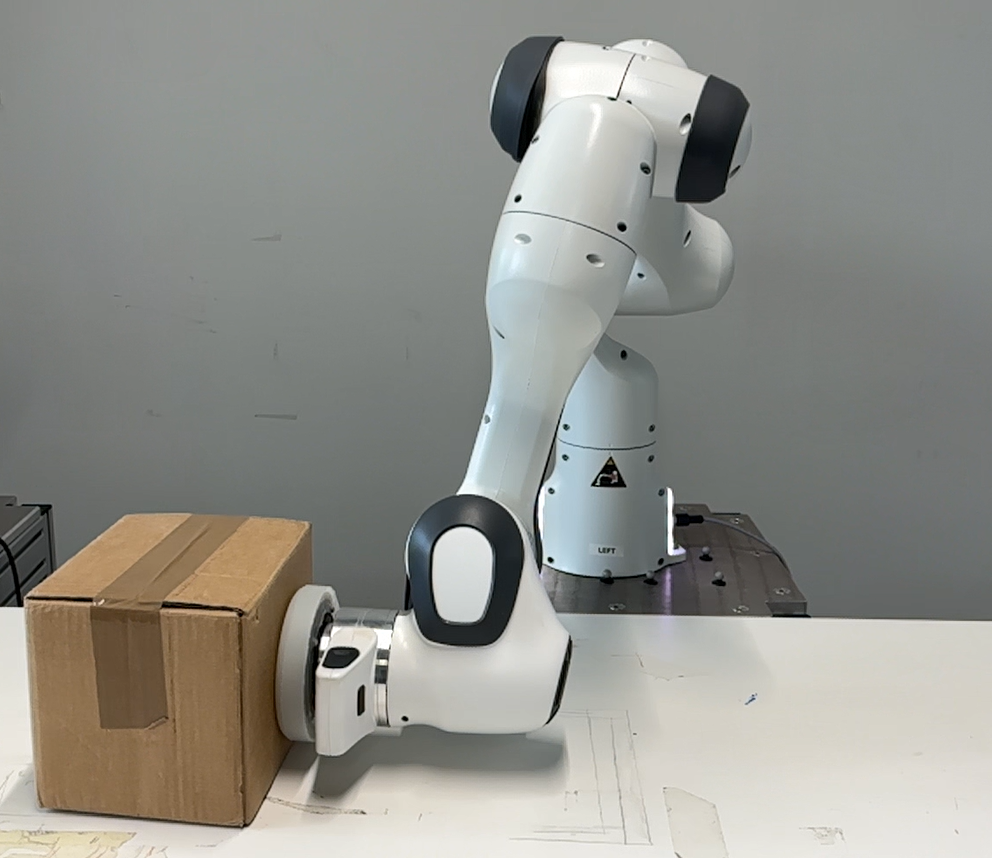}
			\caption{Post-impact state}
			\label{fig:Push_sequence_3}
		\end{subfigure}
		\caption{Snapshots of the system for one of the hit-and-push experiments.}
		\label{fig:snapshots_push}
	\end{figure*}
 
     \begin{figure*}
		\centering
		\begin{subfigure}[b]{0.345\textwidth}
			\centering
			\includegraphics[width=\textwidth]{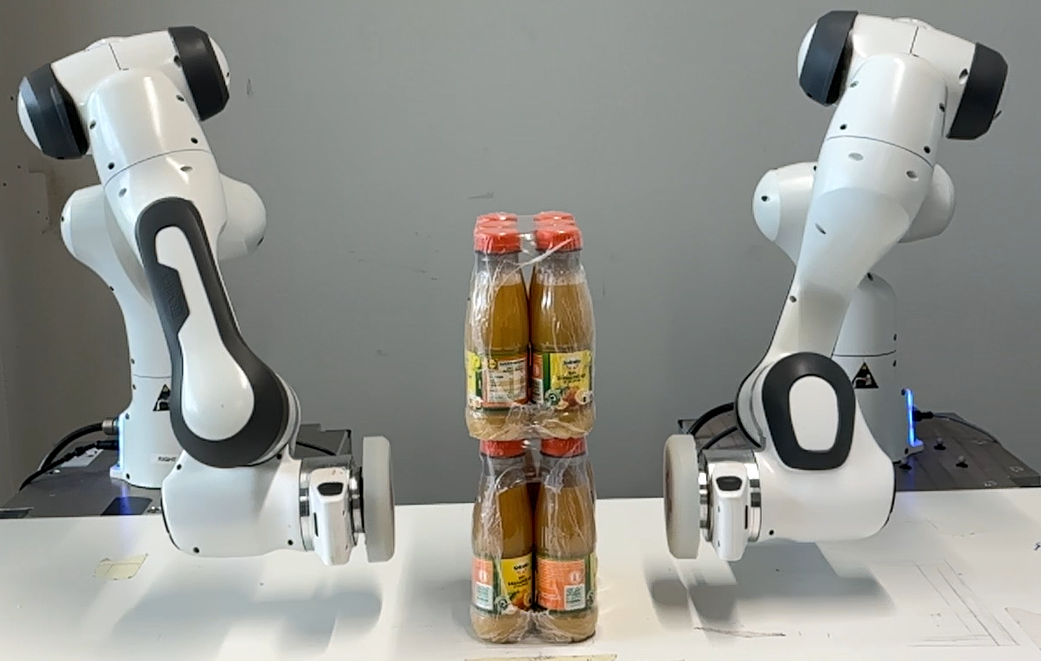}
			\caption{Ante-impact state}
			\label{fig:Grab_sequence_1}
		\end{subfigure}
		\begin{subfigure}[b]{0.325\textwidth}
			\centering
			\includegraphics[width=\textwidth]{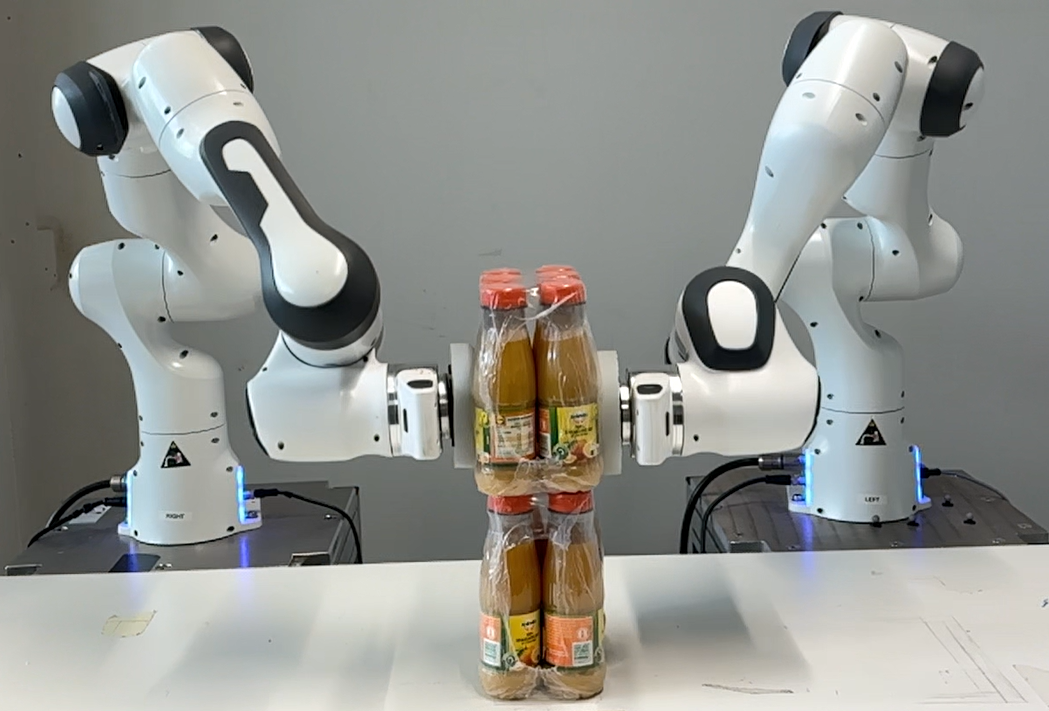}
			\caption{Impact configuration}
			\label{fig:Grab_sequence_2}
		\end{subfigure}
		\begin{subfigure}[b]{0.29\textwidth}
			\centering
			\includegraphics[width=\textwidth]{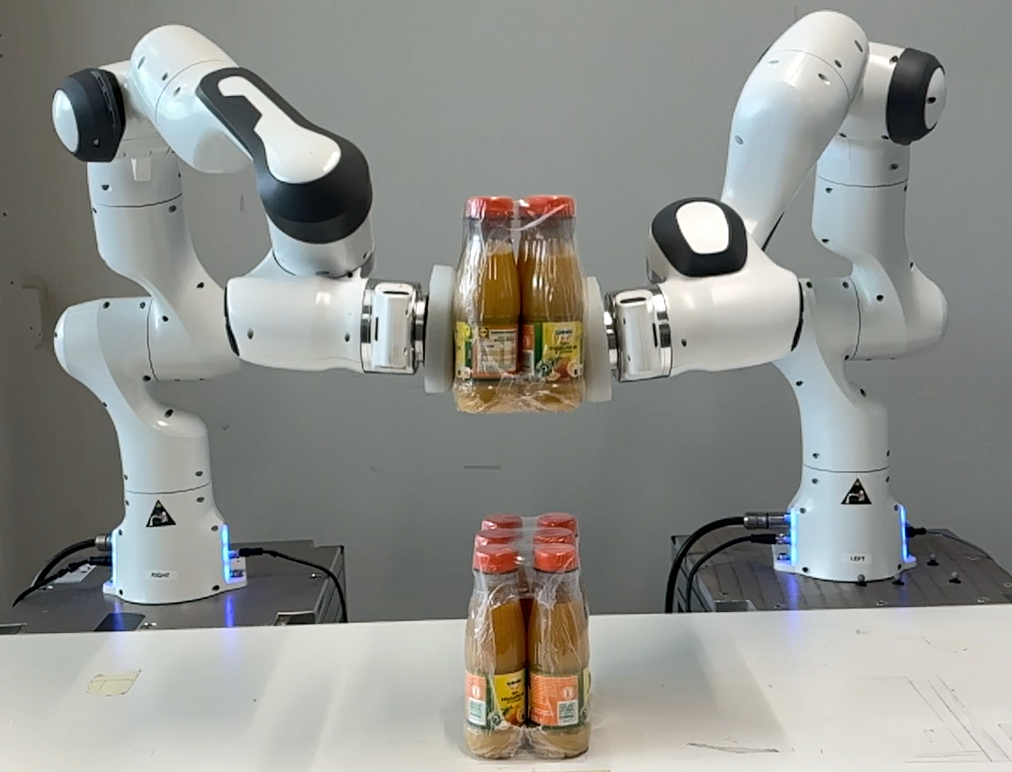}
			\caption{Post-impact state}
			\label{fig:Grab_sequence_3}
		\end{subfigure}
		\caption{Snapshots of the system for one of the grabbing experiments.}
		\label{fig:snapshots_grab}
	\end{figure*}    	   
 
    Instead, we use the property of the RS framework defined in Section \ref{sec:control} that input peaks and jumps are minimized under the assumption that the ante- and post-impact reference comply with the rigid impact map. Combined with acceleration feedforward, friction compensation and compensation of the external force acting the robot, this should lead to a minimum control feedback effort. The core idea of the validation approach is that an inaccurate impact map will imply that the post-impact state of the system does not match the post-impact reference, which will result in a larger feedback effort to work against the natural (impact) dynamics of the system. Hence, the time integral of the feedback signals around the impact time can serve as a measure for the accuracy of the rigid impact map. Since velocity feedback is reduced in the interim mode, as highlighted in Section \ref{sec:control_interim}, the effect of the impact-induced oscillations on the feedback effort is minimized. 
    The feedback signal $\bm a_\text{fb}(t)$ measured during the interim and post-impact mode is defined as
    \begin{equation}\label{eq:fb_error}
    \resizebox{\linewidth}{!}{$
        \bm a_\text{fb}(t) := \left\{   
        \begin{aligned}
        \begin{aligned} 
        2 \sqrt{k_p} & \left( \dot{\bar{\bm p}}^a_{d}(t) - \dot{\bm p}\right) \\ 
       +  k_p & \left( {\bar{\bm p}}^a_{d}(t) - {\bm p}  \right)  
        \end{aligned} \ 
        & 
        \begin{aligned} 
        & \\ & \text{if} \ \  t < T_\text{imp},\end{aligned} \\         
        \begin{aligned} 
          2\sqrt{k_p}\gamma(t)&\left( \dot{\bar{\bm p}}^p_{d}(t) - \dot{\bm p} \right)  \\ 
         +  k_p&\left({\bar{\bm p}}^p_{d}(t) - {\bm p} \right)
        \end{aligned} \ 
        & 
        \begin{aligned} 
        & \\ & \text{if} \ \  T_\text{imp} \leq t \leq T_\text{imp} + \Delta t_\text{int},
          \end{aligned} \\ 
        \begin{aligned} 
        2 \sqrt{k_p} & \left( \dot{\bar{\bm p}}^p_{d}(t) - \dot{\bm p}\right) \\ 
         + k_p & \left( {\bar{\bm p}}^p_{d}(t) - {\bm p}  \right)  
        \end{aligned} \ 
        & 
        \begin{aligned} 
        & \\ & \text{if} \ \  t > T_\text{imp} + \Delta t_\text{int},\end{aligned}
        \end{aligned}\right.
    $}
    \end{equation}
    corresponding to the feedback components of the position task error \eqref{eq:e_pos_ante}, \eqref{eq:e_pos_int} and  \eqref{eq:e_pos_post} for the ante-impact, interim, and post-impact mode, respectively. Note that the feedforward signals, including friction compensation, are not included in \eqref{eq:fb_error} as they contain the predicted nominal control effort that is also present in case of a perfect impact map prediction. The net feedback signal $\bm a_\text{fb,net}$ is then determined by numerically computing the integral of $\bm a_\text{fb}(t)$ over time given by 
    \begin{equation}\label{eq:a_fb_net}
        \bm a_\text{fb,net} = \int_{t_1}^{t_2} \bm a_\text{fb}(t) \mathrm{d}t,
    \end{equation}    
    where $t_1$ is a time before any impact has occurred, and $t_2$ is a time after the impact sequence is completed, and any impact-induced vibrations have vanished. 
    This net feedback signal is then compared against the same signal in experiments where the post-impact reference $\bar{\bm p}^p_{d}(t)$ does not comply with the ante-impact reference $\bar{\bm p}^a_{d}(t)$ through the rigid impact map obtained by the simulator. If applying the reference corresponding to the rigid impact map results in the lowest net feedback signal, this is considered as a validation of the impact map.

    To further clarify why the measure of $\bm a_\text{fb,net}$ is relevant for this validation procedure, we first consider a hypothetical experimental scenario with a rigid joint transmission and instantaneous inelastic impact exactly at the nominal impact time, which would result in an instantaneous velocity jump of the robot(s). Using the proposed control framework, the system would transition into the interim mode at the time of impact detection, followed by the post-impact mode. In both modes, the post-impact reference ${\bar{\bm p}}^p_{d}(t)$ is followed, see \eqref{eq:e_pos_int} and \eqref{eq:e_pos_post}. Because the ante- and post-impact references are compatible with the impact map that is estimated from simulations as described in Section \ref{sec:refgen}, a correct prediction would imply that at an infinitesimal time after the impact impact $T^+_\text{imp}$, it holds that ${\dot{\bar{\bm p}}}^p_{d}(T^+_\text{imp}) = \dot{\bm p}(T^+_\text{imp})$, assuming that the ante-impact reference was perfectly tracked. Because it then also holds that ${\bar{{\bm p}}}^p_{d}(T_\text{imp}) = {\bm p}(T_\text{imp})$ due to \eqref{eq:pos_refs_imp}, the feedback signal $\bm a_\text{fb}(t)$ defined in \eqref{eq:fb_error} will be zero at the time of impact. Because the feedforward signals $\ddot{\bar{\bm p}}^p_{d}(t)$, $\bm a^p_f(t)$ and $\bm a^p_\text{fric}(t)$ compensate for the robot acceleration, external forces and motor friction respectively, the reference ${\bar{\bm p}}^p_{d}(t)$ is continuously followed without requiring any feedback, implying that $\bm a_\text{fb}(t) = 0$ for all $t$, and therefore, $\bm a_\text{fb,net} = 0$. 
    If, however, the predicted impact map is incorrect in this hypothetical rigid scenario, this implies that ${\dot{\bar{\bm p}}}^p_{d}(T^+_\text{imp}) \neq \dot{\bm p}(T^+_\text{imp})$, which will over time result in a position error ${\bar{{\bm p}}}^p_{d}(t) \neq {\bm p}(t)$ on top of a velocity error. This in turn results in a nonzero value of $\bm a_\text{fb}(t)$ over time to catch up with to the reference, and subsequently a nonzero value for $\bm a_\text{fb,net}$. The value of $\bm a_\text{fb,net}$ will increase with a larger mismatch between predicted rigid impact map and the actual velocity jump, making this a good measure for the accuracy of the post-impact map.

    In practice, impacts do not occur instantaneously and impact-induced vibrations are present, which means that ${\dot{\bar{\bm p}}}^p_{d}(T^+_\text{imp}) \neq \dot{\bm p}(T^+_\text{imp})$, even if the rigid impact map is correct. As explained in Section \ref{sec:control_interim}, the interim control mode aims to mitigate the effect of the finite impact duration and impact-induced vibrations on the feedback $\bm a_\text{fb}(t)$ by temporarily reducing the velocity feedback gain after the impact is detected. It will, however, not remove all feedback, implying that a nonzero feedback acceleration $\bm a_\text{fb}(t)$ is commanded on top of the feedforward terms $\ddot{\bar{\bm p}}^p_{d}(t)$, $\bm a^p_f(t)$ and $\bm a^p_\text{fric}(t)$. Once the impact sequence is completed and the impact-induced vibrations have vanished, the nonzero value of $\bm a_\text{fb}(t)$ over will have affected the velocity signal. Assuming that the rigid impact map is correctly predicted, the feedback $\bm a_\text{fb}(t)$ will have introduced an error between ${\dot{\bar{\bm p}}}^p_{d}(t)$ and $\dot{\bm p}(t)$ that then needs to be compensated by a feedback signal $\bm a_\text{fb}(t)$ in opposite direction. If the time integral of $\bm a_\text{fb}(t)$ computed in \eqref{eq:a_fb_net} is zero, this implies that the feedback has had no net effect on the velocity, meaning that the predicted rigid impact map was matching the natural velocity jump of the real robot. If the net feedback $\bm a_\text{fb,net}$ is nonzero, the feedback has had a lasting effect on the velocity, with $\bm a_\text{fb,net} > 0$ implying that the post-impact velocity was underestimated, and $\bm a_\text{fb,net} < 0$ implying that it was overestimated. In conclusion, both in the ideal and real case, a signal $\bm a_\text{fb,net}$ converging to zero is an indication of a correct velocity jump prediction.


\section{Experimental results}\label{sec:experiments}

    The validation approach highlighted in the previous section has been applied for two distinct use cases using the setup depicted in Figure~\ref{fig:setup}. The first is a single-arm \textbf{hit-and-push} scenario, shown in Figure~\ref{fig:snapshots_push}, where an object is approached by the robot with nonzero velocity in negative $y$-direction (expressed in the inertial frame $A$ depicted in Figure~\ref{fig:setup}), aiming to establish sustained contact and push the object to a desired position. The second is a dual-arm \textbf{grabbing} scenario, shown in Figure~\ref{fig:snapshots_grab}, approaching an object with velocity in both lateral ($y$-) and vertical ($z$-)directions, aiming for a quick grasp with a positive post-impact velocity in $z$-direction.

\begin{figure}
		\centering
		\begin{subfigure}[b]{0.372\linewidth}
			\centering
			\includegraphics[width=\textwidth]{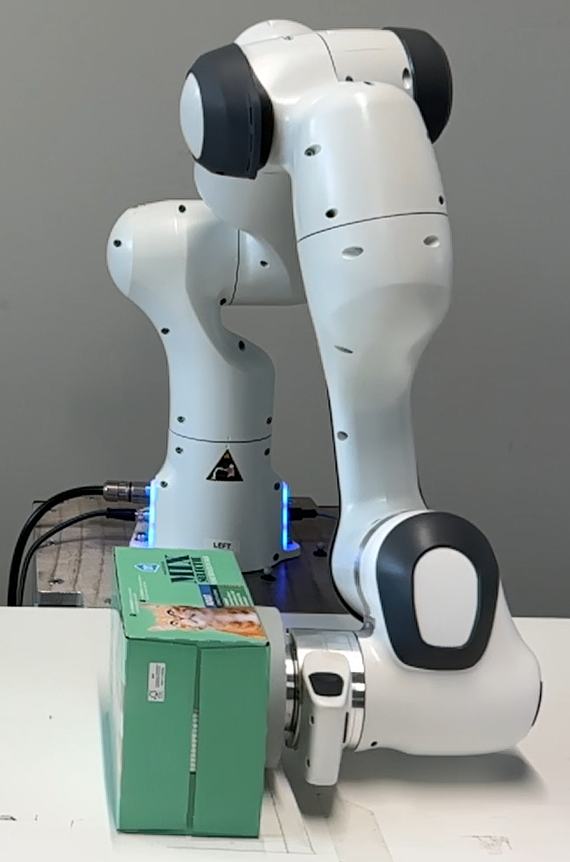}
			\caption{Case 1 \& case 2}
			\label{fig:Push_case_1}
		\end{subfigure}
		\begin{subfigure}[b]{0.36\linewidth}
			\centering
			\includegraphics[width=\textwidth]{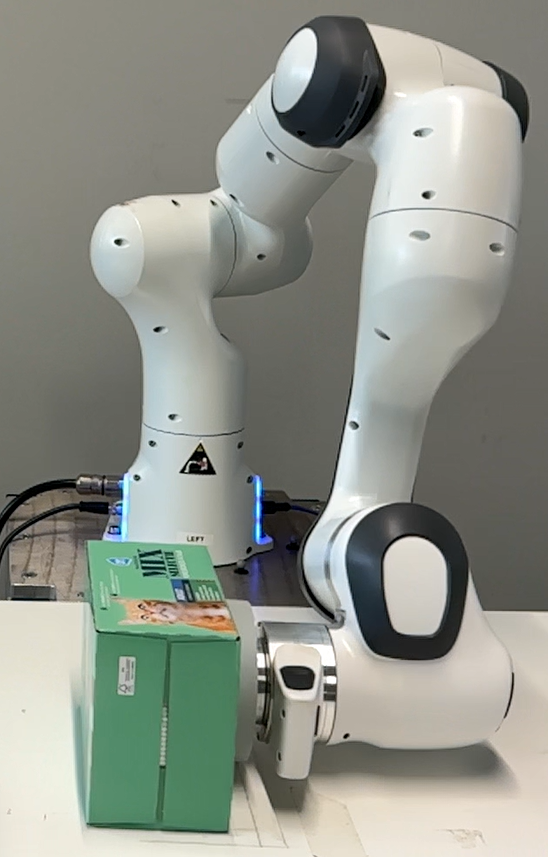}
			\caption{Case 3}
			\label{fig:Push_case_3}
		\end{subfigure}
		\caption{Impact configuration for the set of hit-and-push experiments for the different ante-impact references. All three cases have identical impact position in Cartesian space, but the impact configuration in joint space is modified for case 3 to investigate its effective on the post-impact velocity.}
		\label{fig:null_space_push}
	\end{figure}

    For both scenarios, experiments are performed using three objects of distinct size, shape and material, being 1) a 0.60 kg foam-filled parcel; 2) a 1.30 kg box of cat food; 3) a 2.20 kg pack of 6 juice bottles. Experiments for each use case and object are performed for three different ante-impact states; case 1 prescribes an arbitrary ante-impact state, case 2 prescribes a different ante-impact velocity and identical impact configuration with respect to case 1, and case 3 prescribes an identical ante-impact velocity and position of the end effector in Cartesian space as case 1, but a different impact configuration through a different reference for joint $q_1$, as shown in Figure~\ref{fig:null_space_push}. Case 3 is added to investigate the effect of the null space on the estimated post-impact velocity, as, contrary to standard manipulation established at zero speed, the natural post-impact reference trajectory for the robot depends on the robot configuration at impact time and not just on the end effector pose \cite{Khurana2024}.

    For each case, the corresponding nominal reference is determined using the approach highlighted in Section \ref{sec:refgen}. For each of these three ante-impact references, 5 different post-impact references are evaluated, ranging from 80\% to 120\% of nominal post-impact velocity estimated from the simulation framework. 5 trials are then performed for each post-impact velocity, resulting in a total number of $2 \times 3 \times 3 \times 5 \times 5 = 450$ experimental trials.  
    
    The time integral of $\bm a_\text{fb}$ is then evaluated according to \eqref{eq:a_fb_net} in a time frame ranging from $t_1 = T_\text{nom} - 0.1$ seconds shortly before the nominal impact time to a time $t_2 = T_\text{nom} + 0.6s$ when all post-impact vibrations are definitely settled. Using a linear fit, the \textbf{optimal post-impact velocity} is extracted from the experiments, where optimal implies the best prediction of the rigid post-impact velocity based on the experimental data (i.e. $\bm a_\text{fb,net} = 0$, see \eqref{eq:a_fb_net}). This optimal velocity can be compared to the predicted velocity according to the simulation. 
    
    As additional comparison, a rigid impact map is extracted from simulations for different motor inertia values, swapping the reduced motor inertia $\bm B_\theta$ by: 1) the non-reduced motor inertia $\bm B_\rho$; 2) a zero matrix, indicating only the link inertia is taken into account. This comparison is made to confirm that, also for complex simultaneous impact tasks, using the reduced motor inertia $\bm B_\theta$ as described in Section \ref{sec:eom} in the impact map computation indeed most closely matches reality, as opposed to state-of-the-art approaches that use $\bm B_\rho$ in the impact map calculation such as \cite{Kirner2024}, or no motor inertia such as \cite{Khurana2024}.

\subsection{Single-arm hit-and-push scenario}

\begin{figure}		
		\centering
		\includegraphics[width=\linewidth]{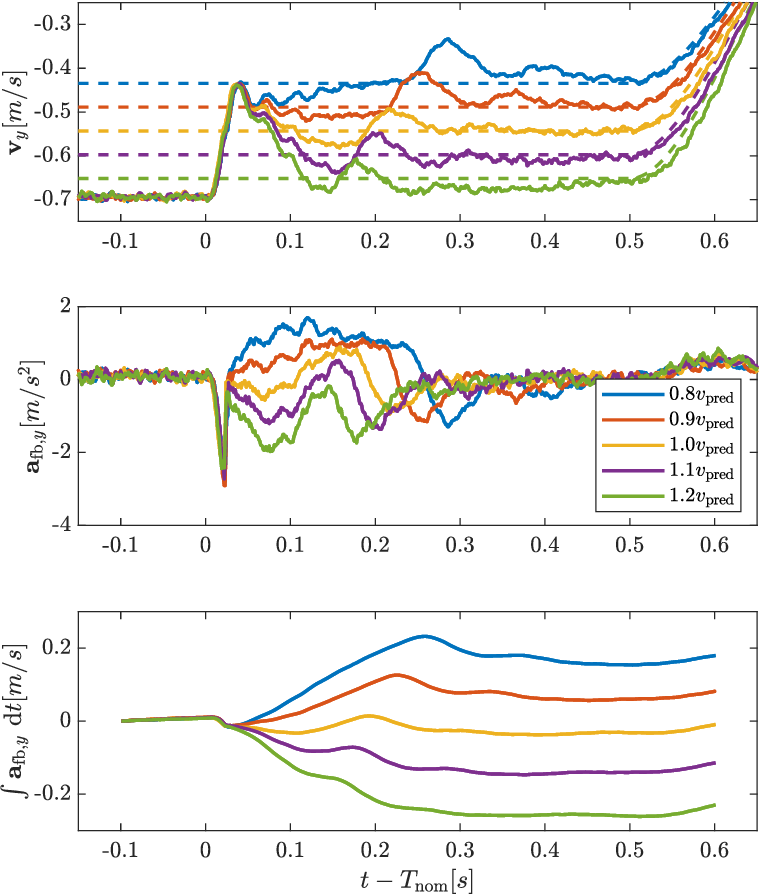}
		\caption{Plots showing the averaged velocity $ v_y$, averaged feedback signal $a_{\text{fb},y}$ and averaged integral of the feedback signal $\int a_{\text{fb},y}$ for 5 different \textbf{hit-and-push} experiments per respective post-impact desired velocity. The legend indicates the post-impact velocity used to generate the reference, expressed as a fraction of the predicted velocity obtained from simulations, indicated as $\bm v_\text{pred}$. The velocity plot additionally shows the desired post-impact velocity references with dashed lines.}
		\label{fig:Single_trial_results_push}
	\end{figure}

    Firstly, the simulation results of a representative hit-and-push experiment set are presented here, using the box with cat food for case 2, which corresponds to an ante-impact velocity of 0.7 m/s.  Figure~\ref{fig:Single_trial_results_push} shows the velocity $v_y$, feedback signal $a_{\text{fb},y}$ and integral of the feedback signal $\int a_{\text{fb},y} \text{d}t$ in the $y$-direction, which is the only relevant direction to consider given the nature of the motion. For each desired post-impact velocity, 5 experiments have been performed, and Figure~\ref{fig:Single_trial_results_push} shows the average value for each signal over these 5 experiments.

    An interesting observation when studying the velocity signal $v_y$ in Figure~\ref{fig:Single_trial_results_push} is that the impact causes a very fast drop in the magnitude of the velocity for all post-impact velocity references, after which the velocity signals start to diverge from each other due to the different post-impact references. This initial drop is larger than the predicted velocity jump corresponding to the rigid impact map, which is due to a combination of impact-induced vibrations in the joints, as well as a temporary loss of contact after the impact, indicating that the impact is partially elastic. At first glance, this seems to contradict the assumption in Section \ref{sec:simulation} that all impacts are inelastic (implying that the coefficient of restitution is 0) in the computation of the rigid impact map. However, due to the constant feedforward signal supplied to the robot in the pushing direction, a series of partially elastic impacts occurs, eventually causing the robot and object to move at the same velocity as if a single inelastic impact has occurred.   
    From the velocity data alone, it is difficult to tell which predicted post-impact velocity is most accurate, as $v_y$ converges to the corresponding desired velocity in all experiments due to velocity feedback. To determine the most accurate rigid impact map, we then have to find out in which experiment the velocity naturally converged to the reference, and in which experiments this conversion is mostly down to the influence of the feedback action as described in Section \ref{sec:validation}. This once again motivates why the validation approach uses the feedback signal as a measure for the accuracy of the prediction,

    
    Considering the feedback signal $a_{\text{fb},y}$ from \eqref{eq:fb_error} in Figure~\ref{fig:Single_trial_results_push}, the first thing to note is a short peak directly after contact is detected. This peak is related to a slight delay in the detection of the contact, and disappears as soon as the impact is detected and the interim mode is entered, which initially removes and then slowly increases velocity feedback to limit the effect on the feedback signal. 
    After this short peak, it becomes clear that the controller actively slows the end effector down if the post-impact velocity is underestimated, and actively speeds up the end effector if the post-impact velocity is overestimated. Note that a positive value of $a_{\text{fb},y}$ corresponds to a deceleration of the end effector, given the negative value of the velocity. 
    
    When the reference corresponds to the predicted post-impact velocity $\bm v_\text{pred}$, the controller initially accelerates the end effector due to the initial velocity drop, but decelerates the end effector roughly a similar amount once the impact-induced vibration vanishes. This is indicated by the small value for the integral of the feedback signal computed according to \eqref{eq:a_fb_net} when all vibrations have vanished, as seen in the bottom plot of Figure~\ref{fig:Single_trial_results_push}. As explained in Section \ref{sec:validation}, the value of this integral at time $t_2 = T_\text{nom} + 0.6 s$ is defined as $\bm a_\text{fb,net}$, which is the measure used to indicate the accuracy of the prediction of the rigid impact map. The closer $\bm a_\text{fb,net}$ is to zero, the more accurate the corresponding predicted rigid impact map is. 
    
\begin{figure}		
		\centering
		\includegraphics[width=\linewidth]{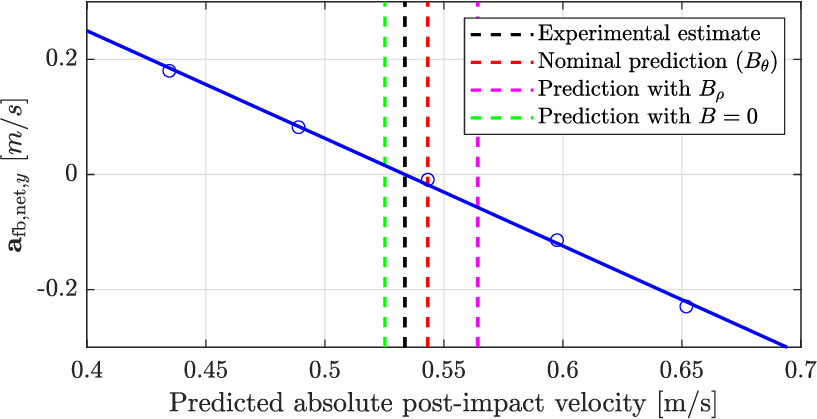}
		\caption{Plot showing a linear fit between the final values of the integrated feedback signals against their corresponding post-impact velocity reference for a \textbf{hit-and-push} motion. The figure additionally shows the experimental estimate of the post-impact velocity using the validation approach, together with the predicted post-impact velocity for simulations with reduced apparent motor inertia $\bm B_\theta$, apparent motor inertia $\bm B_\rho$ and no motor inertia ($\bm B = 0$).}
		\label{fig:Fit_results_push}
	\end{figure}

    Studying the results more closely, we store the value of $\bm a_\text{fb,net}$ for the different experiments in this set, and plot them against the corresponding predicted post-impact velocity, as is done in Figure~\ref{fig:Fit_results_push}. A linear fit through the data points can be made using a least-squares approach to estimate which post-impact velocity intersects with a value of 0 m/s for the integral of the feedback signal. This value is referred to as the \emph{experimental estimate} of the post-impact velocity. We can then determine the percentage mismatch between this experimental estimate and the estimated post-impact velocity, which is indicated as the \emph{nominal prediction}. While this nominal prediction is made using the reduced motor inertia matrix $\bm B_\theta$ according to \eqref{eq:eom}, we attempt to validate this motor inertia model by comparing the nominal prediction against simulations with different inertia models, as mentioned at the start of this section. Simulations using a model with the unreduced motor inertia $\bm B_\rho$ result in a \emph{prediction with $\bm B_\rho$}, and simulations with zero motor inertia give a \emph{prediction with $\bm B = 0$}. 
    
    For this representative set of experiments, we can conclude based on Figure~\ref{fig:Fit_results_push} that the velocity is overestimated by 1.82\% for the nominal prediction and by 5.76\% for the prediction with $\bm B_\rho$, and underestimated by 1.56\% for the prediction with $\bm B = 0$. 
    This would imply that, based on this experiment, one would conclude that the prediction with no motor inertia is roughly as accurate as the nominal prediction. One possible explanation for this is that the value for $\bm B_\theta$ used in these simulations may not be correct, as the actual value of $\bm B_\theta$ for the Franka robots is confidential, as highlighted in Section \ref{sec:eom}. Given the fact that the experimental estimate lies between the prediction using $\bm B_\theta$ and using no motor inertia, it is plausible that the actual confidential value for $\bm B_\theta$ is slightly lower than the value used in this work. In any case, it is clear from these results that the prediction with $\bm B_\rho$ is off by a lot compared to the other two predictions.
    

    As mentioned in the beginning of this section, these experiments are repeated for the three different objects, and three different ante-impact states per object, with case 2 increasing the ante-impact velocity compared to case 1, and case 3 modifying the impact configuration in joint space without changing the references of case 1 in Cartesian space. The percent error between the experimental estimate of the post-impact velocity reference and the predicted post-impact velocities using simulations with different motor inertia values are shown for each case in Table~\ref{tab:comparison_push}. We compute two averaged values, the \textbf{average error} simply averages these percent errors, while the \textbf{absolute error} presents the average of the absolute value of all percent errors, implying that an underestimation in 1 experiment set cannot compensate for an overestimation in another set. The average error represents the bias in the prediction error, while the absolute error represents the average prediction quality.
    
    For the nominal predictions, we see first of all in Table~\ref{tab:comparison_push} that the absolute error is $3.12\%$, which is a reliably small error considering the use cases of this rigid impact map presented in Section \ref{sec:introduction}.  However, we do see a clear average overestimation of the post-impact velocity, indicating that it is indeed plausible that the actual confidential value for $\bm B_\theta$ might be lower than expected. Studying the predictions with different motor inertia, we can see that using no inertia on average results in an underestimation of the impact map, while using $\bm B_\rho$ results in a consistent overestimation, even for every single case.     
    The larger absolute error and clear bias confirm that using $\bm B_\rho$ in the impact map prediction is not accurate, implying that the inertia reduction low-level torque control law of \eqref{eq:torque_control_law} does indeed cause an apparent reduction of the motor inertia, also during impacts, despite the fast nature of these impact motions. However, for the prediction with no motor inertia, we can see that the absolute error is comparable/slightly smaller than that of the nominal prediction.     
    Overall, we can therefore draw the conclusion that there is a relatively small error between the nominal estimated post-impact map and the experimental estimate, as for the post-impact map with no motor inertia, but that the prediction that follows from simulations using $B_\rho$ appears to be off by a lot.

    One final thing to take away from the results in Table~\ref{tab:comparison_push} is the influence of the impact configuration on the velocity prediction, apparent from cases 1 and 3 for each object. Between these cases, only the impact configuration in the joint space is modified, keeping the exact same tasks for the Cartesian pose and velocity. Despite the minor change in impact configuration, shown in Figure~\ref{fig:null_space_push}, the optimal post-impact velocities and the simulation results are significantly affected, especially for the experiment with the pack of juice bottles, confirming that the joint configuration matters for impact-aware manipulation.
    
\begin{table}[]
\caption{Experimental estimates of the post-impact velocity for different hit-and-push experiments, and the percentage mismatch between this experimental estimate and the predicted post-impact velocities from numerical simulations using reduced apparent motor inertia $\bm B_\theta$, apparent motor inertia $\bm B_\rho$ and no motor inertia ($\bm B = 0$). The final rows show the averaged percent error, and the average absolute error, implying that an underestimation in one experiment does not compensate for an overestimation in another experiment.}
\begin{tabular}{c|c|c|c|c}
& \begin{tabular}[c]{@{}c@{}}{\scriptsize Experimental} \\ {\scriptsize estimate}\end{tabular} & \begin{tabular}[c]{@{}c@{}} {\scriptsize Nominal}\\ {\scriptsize prediction}\end{tabular} & \begin{tabular}[c]{@{}c@{}}{\scriptsize Prediction}\\ {\scriptsize with $\bm B_\rho$}\end{tabular} & \begin{tabular}[c]{@{}c@{}}{\scriptsize Prediction}\\ {\scriptsize with $\bm B = 0$}\end{tabular} \\ \hline
{\scriptsize Parcel: case 1 }& $0.425 \ m/s$ & $3.12 \%$ & $5.93 \%$ & $0.25 \%$ \\ \hline 
{\scriptsize Parcel: case 2 }& $0.587 \ m/s$ & $4.36 \%$ & $6.62 \%$ & $2.31 \%$ \\ \hline 
{\scriptsize Parcel: case 3 }& $0.438 \ m/s$ & $0.10 \%$ & $2.90 \%$ & $-2.27 \%$ \\ \hline 
{\scriptsize Cat food: case 1 }& $0.370 \ m/s$ & $4.78 \%$ & $9.52 \%$ & $0.73 \%$ \\ \hline 
{\scriptsize Cat food: case 2 }& $0.534 \ m/s$ & $1.82 \%$ & $5.76 \%$ & $-1.56 \%$ \\ \hline 
{\scriptsize Cat food: case 3 }& $0.380 \ m/s$ & $2.62 \%$ & $6.47 \%$ & $-1.15 \%$ \\ \hline 
{\scriptsize Juice: case 1 }& $0.333 \ m/s$ & $3.19 \%$ & $8.26 \%$ & $-2.98 \%$ \\ \hline 
{\scriptsize Juice: case 2 }& $0.469 \ m/s$ & $2.85 \%$ & $6.77 \%$ & $-1.89 \%$ \\ \hline 
{\scriptsize Juice: case 3 }& $0.364 \ m/s$ & $-5.21 \%$ & $0.31 \%$ & $-10.47 \%$ \\ \hline 
{\scriptsize Average error }& - & $1.96 \%$ & $5.84 \%$ & $-1.89 \%$ \\ \hline 
{\scriptsize Absolute error }& - & $3.12 \%$ & $5.84 \%$ & $2.62 \%$                                     
\end{tabular}    
\label{tab:comparison_push}
\end{table}

\subsection{Dual-arm grabbing scenario}

    As highlighted at the top of this section, a similar set of experiments was performed for a dual-arm grabbing use case. For these experiments, we approach the objects in a symmetrical way with velocities in $y$- and $z$-direction. The relevant quantity to measure the accuracy of the post-impact velocity estimate is the integrated acceleration in $z$-direction, given that the post-impact velocities in $x$- and $y$-direction are nearly equal to zero. As with the hit-and-push experiments, this is done for three objects and three ante-impact states, with 5 corresponding post-impact references, and 5 trials per combination of ante- and post-impact reference. 
    
    \begin{figure}		
		\centering
		\includegraphics[width=\linewidth]{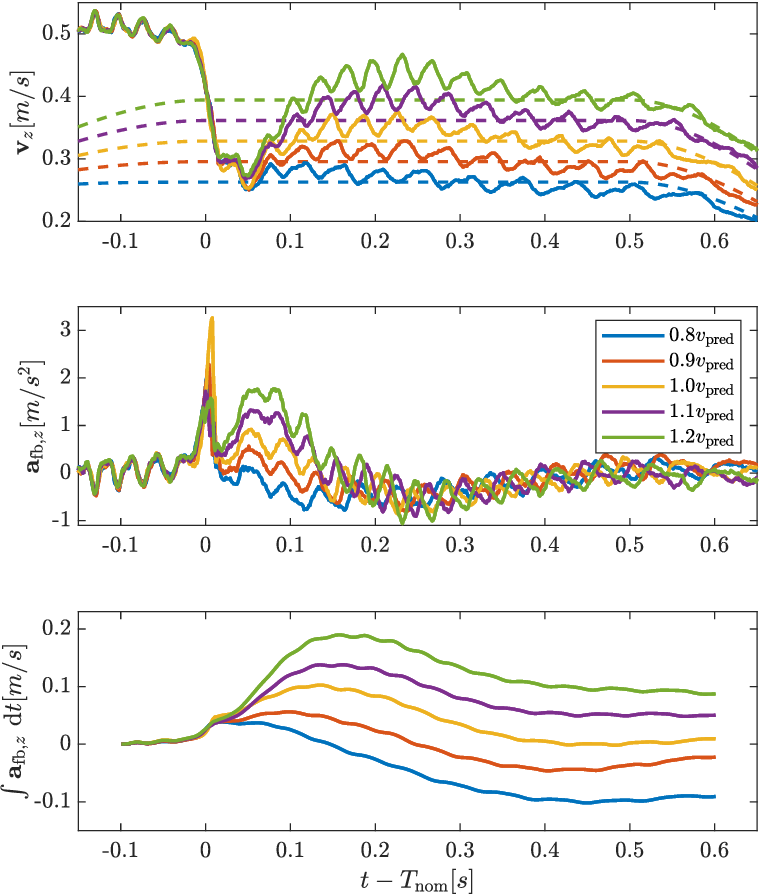}
		\caption{Plots showing the averaged velocity $\bm v_z$, averaged feedback signal $\bm a_{\text{fb},z}$ and averaged integral of the feedback signal $\int \bm a_{\text{fb},z}$ for 5 different \textbf{grabbing} experiments per respective post-impact desired velocity. The legend indicates the post-impact velocity used to generate the reference, expressed as a fraction of the predicted velocity through simulations, indicated as $\bm v_\text{pred}$. The velocity plot additionally shows the desired post-impact velocity references with dashed lines.}
		\label{fig:Single_trial_results_grab}
    \end{figure}
    \begin{figure}		
    \centering
    \includegraphics[width=\linewidth]{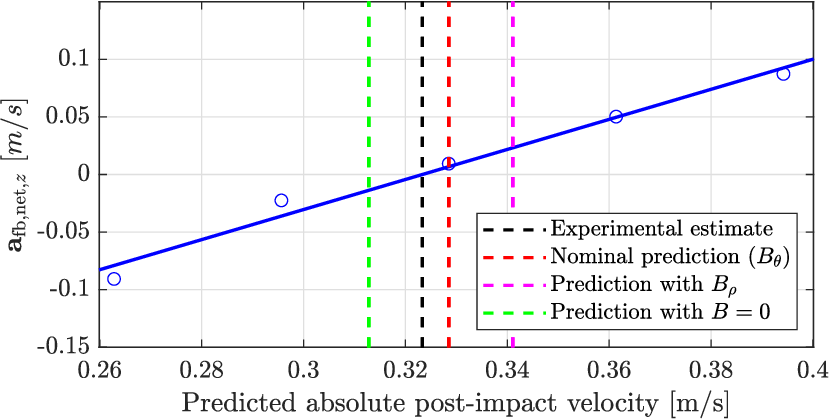}
    \caption{Plot showing a linear fit between the final values of the integrated feedback signals against their corresponding post-impact velocity reference for a \textbf{grabbing} motion. The figure additionally shows the experimental estimate of the post-impact velocity using the validation approach, together with the predicted post-impact velocity for simulations with reduced motor inertia $\bm B_\theta$, motor inertia $\bm B_\rho$ and no motor inertia ($\bm B = 0$).}
    \label{fig:Fit_results_grab}
    \end{figure}
    
    While the framework remains mostly unchanged between the hit-and-push and the grabbing motions, it must be noted that the post-impact orientation control gain is lowered with respect to the ante-impact state, unlike in the hit-and-push case. This is due to the significant effect that the post-impact angular velocity can have upon initial contact for the grabbing use case. While this post-impact angular velocity is ideally zero, a non-negligible angular velocity jump was measured in the experiments, presumably due to the lack of a spherical wrist on the Franka robots used in this work. 
    While the interim mode in the proposed control approach reduces the effect of this angular velocity jump on the control inputs to some extent, the post-impact orientation control gain is additionally lowered for the grabbing experiments upon entering the post-impact state. The clamping force prescribed by the controller then promotes self-alignment, resulting in a successful grasp despite the lower orientation gains. 

    Figure~\ref{fig:Single_trial_results_grab} shows the velocity, feedback signal and integral of the feedback signal for a representative set of grabbing experiments, corresponding to case 3 of the juice bottles. To clearly quantify the results in a single plot despite dealing with dual-arm manipulation, Figure~\ref{fig:Single_trial_results_grab} plots the average velocity $v_z$, input $a_{z,\text{fb}}$ and integral $\int a_{z,\text{fb}}$ between both robots. Despite the different use case, these results paint a very similar image as for the hit-and-push use case, as the velocity initially drops below the predicted post-impact velocity for all cases, and starts converging to the corresponding post-impact velocity afterwards. The oscillations induced by the impact are significantly larger compared to the pushing case, but do die out eventually. While these vibrations are also more apparent in the feedback signal $a_{z,\text{fb}}$, the integral of this signal again shows that the experiment with nominal post-impact velocity reference reaches with a net feedback that is very close to 0 at $t = T_\text{nom} + 0.6 s$ when all vibrations have died out.

 \begin{table}[]
\caption{Experimental estimates of the post-impact velocity for different grabbing experiments, compared against the predicted post-impact velocity from numerical simulations using reduced apparent motor inertia $\bm B_\theta$, apparent motor inertia $\bm B_\rho$ and no motor inertia ($\bm B = 0$).}
\label{tab:comparison_grab}
\begin{tabular}{c|c|c|c|c}
& \begin{tabular}[c]{@{}c@{}}{\scriptsize Experimental} \\ {\scriptsize estimate}\end{tabular} & \begin{tabular}[c]{@{}c@{}} {\scriptsize Nominal}\\ {\scriptsize prediction}\end{tabular} & \begin{tabular}[c]{@{}c@{}}{\scriptsize Prediction}\\ {\scriptsize with $\bm B_\rho$}\end{tabular} & \begin{tabular}[c]{@{}c@{}}{\scriptsize Prediction}\\ {\scriptsize with $\bm B = 0$}\end{tabular} \\ \hline
{\scriptsize Parcel: case 1} & $0.361 \ m/s$ & $4.04 \%$ & $7.42 \%$ & $3.67 \%$ \\ \hline 
{\scriptsize Parcel: case 2} & $0.505 \ m/s$ & $11.06 \%$ & $14.08 \%$ & $9.82 \%$ \\ \hline 
{\scriptsize Parcel: case 3} & $0.379 \ m/s$ & $1.86 \%$ & $3.73 \%$ & $1.40 \%$ \\ \hline 
{\scriptsize Cat food: case 1} & $0.336 \ m/s$ & $1.18 \%$ & $5.35 \%$ & $-1.46 \%$ \\ \hline 
{\scriptsize Cat food: case 2} & $0.499 \ m/s$ & $2.66 \%$ & $6.06 \%$ & $-3.41 \%$ \\ \hline 
{\scriptsize Cat food: case 3} & $0.372 \ m/s$ & $-4.62 \%$ & $-3.81 \%$ & $-4.80 \%$ \\ \hline 
{\scriptsize Juice: case 1} & $0.316 \ m/s$ & $1.15 \%$ & $8.03 \%$ & $-7.82 \%$ \\ \hline 
{\scriptsize Juice: case 2} & $0.516 \ m/s$ & $-7.56 \%$ & $-1.77 \%$ & $-17.31 \%$ \\ \hline 
{\scriptsize Juice: case 3} & $0.323 \ m/s$ & $1.62 \%$ & $5.50 \%$ & $-3.24 \%$ \\ \hline 
{\scriptsize Average error} & - & $1.27 \%$ & $4.95 \%$ & $-2.57 \%$ \\ \hline 
{\scriptsize Absolute error} & - & $3.97 \%$ & $6.19 \%$ & $5.88 \%$ 
\end{tabular}  
\end{table}
    
    The integral values at $t = T_\text{nom} + 0.6 s$, called $\bm a_\text{fb,net}$, are plotted and fitted as done for the hit-and-push motion to determine the experimental estimate of the post-impact velocity, shown in Figure~\ref{fig:Fit_results_grab}. As with the hit-and-push case, the nominal prediction slightly overestimates the predicted post-impact velocity, but the error is quite small.     
    Despite this slight overestimation of the post-impact velocity for the nominal prediction, the prediction error is smaller than the prediction error when using $\bm B_\rho$ or no motor inertia ($\bm B = 0)$, indicating that the nominal prediction, which uses $\bm B_\theta$, is the best available model for this case. 

    Table~\ref{tab:comparison_grab} shows the experimentally estimated post-impact velocities with the prediction errors for all grabbing experiment sets, and from these results, one can indeed conclude that the nominal prediction is accurate with an average absolute error of 3.12\%, similar to that in the hit-and-push case. However, for the grabbing case, the nominal prediction is also the most accurate one, as both the average error and the absolute error are the smallest for this nominal prediction when compared to the predictions using different motor inertia models. Taking the combined average of the hit-and-push and grabbing experiments, one can draw the conclusion that the nominal prediction, which uses a model with motor inertia $\bm B_\theta$, is the best available prediction of the rigid impact map with an average prediction error of $3.55\%$ against $6.02 \%$ for a model with $\bm B_\rho$ and $4.25\%$ for a model with no motor inertia. While the average overestimation of the post-impact velocity indicates that the exact confidential values for $\bm B_\theta$ are likely lower than those presented in Table~\ref{tab:parameters_B}, it is therefore still the most accurate model available.
    

\section{Conclusion}\label{sec:conclusion}

    This work has presented a framework for experimentally validating the accuracy of a rigid impact map that is obtained from a physics engine, which indicates the gross velocity jump without modeling impact-induced vibrations. The approach uses the reference spreading control framework to create a reference that is consistent with the impact dynamics by extracting the rigid impact map from a physics engine. Following the logic that a correct estimation of the impact map should result in an integral of the feedback signal equal to zero, the rigid impact map is estimated from experiments with different post-impact velocity references. A comparison of this experimentally estimated rigid impact map with a rigid impact map predicted using nonsmooth simulations shows an average absolute error of 3.55\% in the post-impact velocity. This nominal prediction uses simulations with a reduced motor inertia according to the low-level torque control law. A comparison with simulations with the unreduced motor inertia or no motor inertia results in an average absolute error of 6.02\% and 4.25\%, respectively, highlighting that the model with reduced motor inertia is indeed the best available model. The lack of input peaks in the results with the correct velocity prediction simultaneously validates of the applicability of reference spreading when using a physics engine in the reference generation procedure.

    The average estimation error of 3.55\% can partially be attributed to the fact that the exact motor inertia values and low-level torque control gains for the Franka robots are unknown. A possible future extension would be to perform these experiments in collaboration with the manufacturer, as was done in \cite{Arias2024}, to evaluate if the resulting estimation error becomes even smaller. Furthermore, the presence of a significant post-impact angular velocity jump can also negatively influence the validity of the results. Given the fact that such a jump is to be avoided for successful execution of the presented impact-aware tasks, a relevant future endeavor could be an extensive analysis of different robot designs regarding the suitability for impact-aware manipulation.

\addtolength{\textheight}{-0cm}  
	
\section*{APPENDIX}

\subsection{Friction compensation}\label{app:friction_compensation}

    Unless properly compensated, the effects of joint friction torques $\bm \tau_\text{fric}$ are non-negligible in torque controlled robots. In practice, we experience that common approaches such as \cite{Gaz2019} underestimate the level of friction, leaving friction to be partially compensated by feedback. Since the validation approach relies on an evaluation of such feedback signals as explained in Section \ref{sec:validation}, it is important that friction is very accurately compensated by acceleration feedforward terms $\bm a^a_\text{fric}(t)$ and $\bm a^p_\text{fric}(t)$. This is done through separate friction identification experiments for the ante- and post-impact control mode described in Section \ref{sec:control_ante} and Section \ref{sec:contol_post}, with friction $\bm a^a_\text{fric}(t)$ and $\bm a^p_\text{fric}(t)$ set to zero. The lack of friction compensation will be compensated by an increase of the feedback, as indicated by $\bm a^a_{i,\text{fric}}(t)$ for each experiment $i$ according to
    \begin{equation}
        \bm a^a_{i,\text{fric}}(t) := 2\sqrt{k_p}\left( \dot{\bar{\bm p}}^a_{d}(t) - \dot{\bm p}_i(t) \right) + k_p\left( {\bar{\bm p}}^a_{d}(t) - {\bm p}_i(t) \right)
    \end{equation}
    with index $i\in N^a_\text{fric}$. The average over this total number of $N^a_\text{fric}$ experiments is then taken to give 
    \begin{equation}
        \bm a^a_\text{fric}(t) := \frac{1}{N^a_\text{fric}} \sum_{i=1}^{N^a_\text{fric}} \bm a^a_{i,\text{fric}}(t).
    \end{equation}
    A similar approach is followed using the post-impact QP to determine $\bm a^p_\text{fric}(t)$.
	
\bibliography{References/library}{}

\begin{thebibliography}{10}

\bibitem{Zhang2020}
C.~Zhang, W.~Zou, L.~Ma, and Z.~Wang, ``{Biologically inspired jumping robots:
  A comprehensive review},'' {\em Robotics and Autonomous Systems}, vol.~124,
  no.~103362, 2020.

\bibitem{Katz2019}
B.~Katz, J.~D. Carlo, and S.~Kim, ``{Mini cheetah: A platform for pushing the
  limits of dynamic quadruped control},'' in {\em International Conference on
  Robotics and Automation}, pp.~6295--6301, IEEE, 2019.

\bibitem{Wensing2013}
P.~M. Wensing and D.~E. Orin, ``{High-speed humanoid running through control
  with a 3D-SLIP model},'' in {\em International Conference on Intelligent
  Robots and Systems}, pp.~5134--5140, IEEE, 2013.

\bibitem{Khurana2021}
H.~Khurana, M.~Bombile, and A.~Billard, ``{Learning to Hit: A statistical
  Dynamical System based approach},'' in {\em International Conference on
  Intelligent Robots and Systems}, pp.~9415--9421, IEEE, 2021.

\bibitem{Yan2024}
L.~Yan, T.~Stouraitis, {Moura, Joao}, W.~Xu, M.~Gienger, and S.~Vijayakumar,
  ``{Impact-Aware Bimanual Catching of Large-Momentum Objects},'' {\em IEEE
  Transactions on Robotics}, 2024.

\bibitem{Bombile2022}
M.~Bombile and A.~Billard, ``{Dual-Arm Control for Coordinated Fast Grabbing
  and Tossing of an Object: Proposing a New Approach},'' {\em IEEE Robotics and
  Automation Magazine}, vol.~29, no.~3, pp.~127--138, 2022.

\bibitem{Dehio2022}
N.~Dehio, Y.~Wang, and A.~Kheddar, ``{Dual-Arm Box Grabbing with Impact-Aware
  MPC Utilizing Soft Deformable End-Effector Pads},'' {\em IEEE Robotics and
  Automation Letters}, vol.~7, no.~2, pp.~5647--5654, 2022.

\bibitem{Zermane2024}
A.~Zermane, N.~Dehio, and A.~Kheddar, ``{Planning Impact-Driven Logistic
  Tasks},'' {\em IEEE Robotics and Automation Letters}, vol.~9, no.~3,
  pp.~2184--2191, 2024.

\bibitem{Haddadin2017}
S.~Haddadin, A.~{De Luca}, and A.~Albu-Sch{\"{a}}ffer, ``{Robot collisions: A
  survey on detection, isolation, and identification},'' {\em IEEE Transactions
  on Robotics}, vol.~33, no.~6, pp.~1292--1312, 2017.

\bibitem{Wang2019}
Y.~Wang and A.~Kheddar, ``{Impact-Friendly Robust Control Design with
  Task-Space Quadratic Optimization},'' in {\em Proceedings of Robotics:
  Science and Systems}, 2019.

\bibitem{Kirner2024}
A.~Kirner and C.~Ott, ``{Impact Analysis for the Planning of Targeted
  Non-Slippage Impacts of Robot Manipulators},'' {\em IEEE Robotics and
  Automation Letters}, vol.~9, no.~3, pp.~2750--2757, 2024.

\bibitem{Khurana2024}
H.~Khurana and A.~Billard, ``{Motion Planning and Inertia-Based Control for
  Impact Aware Manipulation},'' {\em IEEE Transactions on Robotics}, vol.~40,
  pp.~2201--2216, 2024.

\bibitem{Proper2023}
B.~Proper, A.~Kurdas, S.~Abdolshah, S.~Haddadin, and A.~Saccon, ``{Aim-Aware
  Collision Monitoring: Discriminating Between Expected and Unexpected
  Post-Impact Behaviors},'' {\em IEEE Robotics and Automation Letters}, vol.~8,
  no.~8, pp.~4609--4616, 2023.

\bibitem{Yang2021}
W.~Yang and M.~Posa, ``{Impact Invariant Control with Applications to Bipedal
  Locomotion},'' in {\em International Conference on Intelligent Robots and
  Systems (IROS)}, pp.~5128--5135, IEEE, 2021.

\bibitem{Steen2024}
J.~van Steen, G.~van~den Brandt, N.~van~de Wouw, J.~Kober, and A.~Saccon,
  ``{Quadratic Programming-based Reference Spreading Control for Dual-Arm
  Robotic Manipulation with Planned Simultaneous Impacts},'' {\em IEEE
  Transactions on Robotics}, vol.~40, pp.~3341--3355, 2024.

\bibitem{Aouaj2021}
I.~Aouaj, V.~Padois, and A.~Saccon, ``{Predicting the Post-Impact Velocity of a
  Robotic Arm via Rigid Multibody Models : an Experimental Study},'' in {\em
  International Conference on Robotics and Automation}, pp.~2264--2271, IEEE,
  2021.

\bibitem{Brogliato2016}
B.~Brogliato, {\em {Nonsmooth Mechanics}}.
\newblock Communications and Control Engineering, Springer International
  Publishing, 3rd~ed., 2016.

\bibitem{Glocker2006}
C.~Glocker, ``{An Introduction to Impacts},'' in {\em Nonsmooth Mechanics of
  Solids, CISM Courses and Lectures}, vol.~485, pp.~45--101, Springer, Vienna,
  2006.

\bibitem{Rijnen2019}
M.~Rijnen, H.~L. Chen, N.~van~de Wouw, A.~Saccon, and H.~Nijmeijer,
  ``{Sensitivity analysis for trajectories of nonsmooth mechanical systems with
  simultaneous impacts: A hybrid systems perspective},'' in {\em American
  Control Conference}, pp.~3623--3629, IEEE, 2019.

\bibitem{Steen2022b}
J.~J. van Steen, A.~Coşgun, N.~van~de Wouw, and A.~Saccon, ``{Dual Arm
  Impact-Aware Grasping through Time-Invariant Reference Spreading Control},''
  {\em IFAC-PapersOnLine}, vol.~56, no.~2, pp.~1009--1016, 2023.

\bibitem{Acary2008}
V.~Acary and B.~Brogliato, {\em {Numerical Methods for Nonsmooth Dynamical
  Systems}}.
\newblock Springer Berlin, Heidelberg, 2008.

\bibitem{Chatterjee1999}
A.~Chatterjee, ``{On the realism of complementarity conditions in rigid body
  collisions},'' {\em Nonlinear Dynamics}, vol.~20, no.~2, pp.~159--168, 1999.

\bibitem{Arias2024}
C.~A.~R. Arias, W.~Weekers, M.~Morganti, V.~Padois, and A.~Saccon, ``{Refined
  Post-Impact Velocity Prediction for Torque-Controlled Flexible-Joint
  Robots},'' {\em IEEE Robotics and Automation Letters}, vol.~9, no.~4,
  pp.~3267--3274, 2024.

\bibitem{Acosta2022}
B.~Acosta, W.~Yang, and M.~Posa, ``{Validating Robotics Simulators on
  Real-World Impacts},'' {\em IEEE Robotics and Automation Letters}, vol.~7,
  pp.~6471--6478, jul 2022.

\bibitem{Jongeneel2023}
M.~J. Jongeneel, L.~Poort, N.~van~de Wouw, and A.~Saccon, ``{Experimental
  Validation of Nonsmooth Dynamics Simulations for Robotic Tossing involving
  Friction and Impacts},'' {\em Preprint: https://hal.science/hal-03974604/},
  2023.

\bibitem{Saccon2014}
A.~Saccon, N.~van~de Wouw, and H.~Nijmeijer, ``{Sensitivity analysis of hybrid
  systems with state jumps with application to trajectory tracking},'' in {\em
  Conference on Decision and Control}, pp.~3065--3070, IEEE, 2014.

\bibitem{Rijnen2015}
M.~Rijnen, A.~Saccon, and H.~Nijmeijer, ``{On Optimal Trajectory Tracking for
  Mechanical Systems with Unilateral Constraints},'' in {\em Conference on
  Decision and Control}, pp.~2561--2566, IEEE, 2015.

\bibitem{Rijnen2017}
M.~Rijnen, E.~{De Mooij}, S.~Traversaro, F.~Nori, N.~van~de Wouw, A.~Saccon,
  and H.~Nijmeijer, ``{Control of humanoid robot motions with impacts:
  Numerical experiments with reference spreading control},'' in {\em
  Proceedings - IEEE International Conference on Robotics and Automation},
  pp.~4102--4107, 2017.

\bibitem{Biemond2013}
J.~J. Biemond, N.~van~de Wouw, W.~P.~H. Heemels, and H.~Nijmeijer, ``{Tracking
  Control for Hybrid Systems With State-Triggered Jumps},'' {\em IEEE
  Transactions on Automatic Control}, vol.~58, no.~4, pp.~876--890, 2013.

\bibitem{Forni2013}
F.~Forni, A.~R. Teel, and L.~Zaccarian, ``{Follow the bouncing ball: Global
  results on tracking and state estimation with impacts},'' {\em IEEE
  Transactions on Automatic Control}, vol.~58, no.~6, pp.~1470--1485, 2013.

\bibitem{Leine2008}
R.~I. Leine and N.~van~de Wouw, {\em {Stability and Convergence of Mechanical
  Systems with Unilateral Constraints}}, vol.~36 of {\em Lecture Notes in
  Applied and Computational Mechanics}.
\newblock Springer Berlin Heidelberg, 2008.

\bibitem{Steen2022}
J.~J. van Steen, N.~van~de Wouw, and A.~Saccon, ``{Robot Control for
  Simultaneous Impact Tasks via Quadratic Programming Based Reference
  Spreading},'' in {\em American Control Conference}, pp.~3865--3872, IEEE,
  2022.

\bibitem{Traversaro2019}
S.~Traversaro and A.~Saccon, ``{Multibody dynamics notation (version 2)},''
  tech. rep., Technische Universiteit Eindhoven, 2019.

\bibitem{Albu2007}
A.~Albu-Sch{\"{a}}ffer, C.~Ott, and G.~Hirzinger, ``{A Unified Passivity-based
  Control Framework for Position, Torque and Impedance Control of Flexible
  Joint Robots},'' {\em The International Journal of Robotics Research},
  vol.~26, no.~1, pp.~23--39, 2007.

\bibitem{Ott2002}
C.~Ott, A.~Albu-Sch{\"{a}}ffer, and G.~Hirzinger, ``{Comparison of adaptive and
  nonadaptive tracking control laws for a flexible joint manipulator},'' in
  {\em IEEE International Conference on Intelligent Robots and Systems},
  pp.~2018--2024, 2002.

\bibitem{Gaz2019}
C.~Gaz, M.~Cognetti, A.~Oliva, P.~R. Giordano, and A.~de~Luca, ``{Dynamic
  identification of the Franka Emika Panda Robot with retrieval of feasible
  parameters using penalty-based optimization},'' {\em IEEE Robotics and
  Automation Letters}, vol.~4, no.~4, pp.~4147--4154, 2019.

\bibitem{Wiberg2022}
V.~Wiberg, E.~Wallin, T.~Nordfjell, and M.~Servin, ``{Control of Rough Terrain
  Vehicles Using Deep Reinforcement Learning},'' {\em IEEE Robotics and
  Automation Letters}, vol.~7, no.~1, pp.~390--397, 2022.

\bibitem{Li2020}
G.~Li, H.~B. Waldum, M.~O. Grindvik, R.~S. J{\o}rundl, and H.~Zhang,
  ``{Development of a vision-based target exploration system for snake-like
  robots in structured environments},'' {\em International Journal of Advanced
  Robotic Systems}, vol.~17, no.~4, 2020.

\bibitem{Styrud2022}
J.~Styrud, M.~Iovino, M.~Norrlof, M.~Bjorkman, and C.~Smith, ``{Combining
  Planning and Learning of Behavior Trees for Robotic Assembly},'' in {\em IEEE
  International Conference on Robotics and Automation}, pp.~11511--11517, 2022.

\bibitem{Cao2023}
M.~Cao, K.~Cao, S.~Yuan, T.~M. Nguyen, and L.~Xie, ``{NEPTUNE: Nonentangling
  Trajectory Planning for Multiple Tethered Unmanned Vehicles},'' {\em IEEE
  Transactions on Robotics}, vol.~39, no.~4, pp.~2786--2804, 2023.

\bibitem{Omer2024}
M.~Omer, K.~Merckaert, and B.~Vanderborght, ``{Constraint control for
  non-prehensile robotic transportation},'' in {\em 43rd Benelux Meeting on
  Systems and Control}, p.~82, 2024.

\bibitem{Lacoursiere2007}
C.~Lacoursi{\`{e}}re, {\em {Ghosts and machines: regularized variational
  methods for interactive simulations of multibodies with dry frictional
  contacts}}.
\newblock PhD thesis, Ume{\aa} University, 2007.

\bibitem{Khatib1987}
O.~Khatib, ``{A unified approach for motion and force control of robot
  manipulators: The operational space formulation},'' {\em IEEE Journal on
  Robotics and Automation}, vol.~3, pp.~43--53, feb 1987.

\bibitem{Spong2020}
M.~W. Spong, S.~Hutchinson, and M.~Vidyasagar, ``{Path and Trajectory
  Planning},'' in {\em Robot Modeling and Control}, ch.~7, pp.~215--268, Wiley,
  2~ed., 2020.

\bibitem{Bouyarmane2019}
K.~Bouyarmane, K.~Chappellet, J.~Vaillant, and A.~Kheddar, ``{Quadratic
  Programming for Multirobot and Task-Space Force Control},'' {\em IEEE
  Transactions on Robotics}, vol.~35, no.~1, pp.~64--77, 2019.

\bibitem{Salini2010}
J.~Salini, S.~Barth{\'{e}}lemy, and P.~Bidaud, ``{LQP-Based Controller Design
  for Humanoid Whole-Body Motion},'' in {\em Advances in Robot Kinematics:
  Motion in Man and Machine}, pp.~177--184, Springer, 2010.

\end{thebibliography}
\bibliographystyle{ieeetr}

\end{document}